\RequirePackage{fix-cm}
\documentclass[smallcondensed]{svjour3}     
\smartqed  
\usepackage{graphicx}
\usepackage{makeidx}         
\DeclareGraphicsExtensions{.png,.pdf}
\usepackage{color}
\usepackage{commath}
\usepackage{amssymb}
\usepackage{amsmath}
\usepackage{amsfonts}
\usepackage{times}
\usepackage{float}
\usepackage{mathtools,xparse}
\usepackage{algorithm}
\usepackage{algpseudocode}
\usepackage{array}
\usepackage{colortbl}
\usepackage[utf8]{inputenc}
\usepackage[T1]{fontenc}
\usepackage{dsfont}
\usepackage{etoolbox}
\usepackage{lipsum}
\usepackage{afterpage}
\usepackage[makeroom]{cancel}
\usepackage{cite}

\usepackage{nomencl}
\makenomenclature

\DeclareMathOperator*{\argmax}{argmax} 
\DeclareMathOperator*{\argmin}{argmin}

\newcolumntype{M}[1]{>{\centering\arraybackslash}m{#1}}

\begin{document}

\title{DARE-SLAM: Degeneracy-Aware and Resilient Loop Closing in Perceptually-Degraded Environments
\thanks{The research was carried out at the Jet Propulsion Laboratory, California Institute of Technology, under a contract with the National Aeronautics and Space Administration. This work was partially funded by the Defense Advanced Research Projects Agency (DARPA). Government Sponsorship acknowledged. \textcopyright  2020 All rights reserved.}
}
\author{Kamak Ebadi, Matteo Palieri, Sally Wood, Curtis Padgett, Ali-akbar Agha-mohammadi}

\institute{Kamak Ebadi \at
              NASA Jet Propulsion Laboratory, California Institute of Technology - Santa Clara University \\
              \email{ebadi@jpl.nasa.gov}
           \and
           Matteo Palieri \at
              NASA Jet Propulsion Laboratory, California Institute of Technology - Polytechnic University of Bari      
           \and
           Sally Wood \at
              Department of Electrical and Computer Engineering - Santa Clara University
            \and
           Curtis Padgett, and Ali-akbar Agha-mohammadi \at
              NASA Jet Propulsion Laboratory - California Institute of Technology
}

\maketitle

\begin{abstract}
Enabling fully autonomous robots capable of navigating and exploring large-scale, unknown and complex environments has been at the core of robotics research for several decades.
A key requirement in autonomous exploration is building accurate and consistent maps of the unknown environment that can be used for reliable navigation.
Loop closure detection, the ability to assert that a robot has returned to a previously visited location, is crucial for consistent mapping as it reduces the drift caused by error accumulation in the estimated robot trajectory. Moreover, in multi-robot systems, loop closures enable merging local maps obtained by a team of robots into a consistent global map of the environment.
In this paper, we present a degeneracy-aware and drift-resilient loop closing method to improve place recognition and resolve 3D location ambiguities for simultaneous localization and mapping (SLAM) in GPS-denied, large-scale and perceptually-degraded environments. More specifically, we focus on SLAM in subterranean environments (e.g., lava tubes, caves, and mines) that represent examples of complex and ambiguous environments where current methods have inadequate performance.
The first contribution of this paper is a degeneracy-aware lidar-based SLAM front-end to determine the observability and level of geometric degeneracy in an unknown environment. Using this crucial capability, ambiguous and unobservable areas in an unknown environment are determined and excluded from the search for loop closures to avoid distortions of the entire map as the result of spurious or inaccurate loop closures.
The second contribution of this paper is a drift-resilient loop closing pipeline that exploits the salient 2D and 3D features extracted from lidar point cloud data to enable a robust multi-stage loop closing capability. Unlike methods that perform the search for loop closures locally to compensate for the ambiguity and high computational cost associated with lidar-based loop closures in large-scale environments, our proposed method relies on a rapid pre-matching step that enables the search over the entire robot trajectory, and thus, it is not affected by drift and accumulation of errors in robot trajectory.
We present extensive evaluation and analysis of performance and robustness, and provide comparison of localization and mapping results with the state-of-the-art methods in a variety of extreme and perceptually-degraded underground mines across the United States.
\end{abstract}

\section{Introduction} \label{sec:introduction}
\begin{figure}[b!]
\centering
	\includegraphics[width=1.0\columnwidth, trim= 0mm 0mm 0mm 0mm, clip]{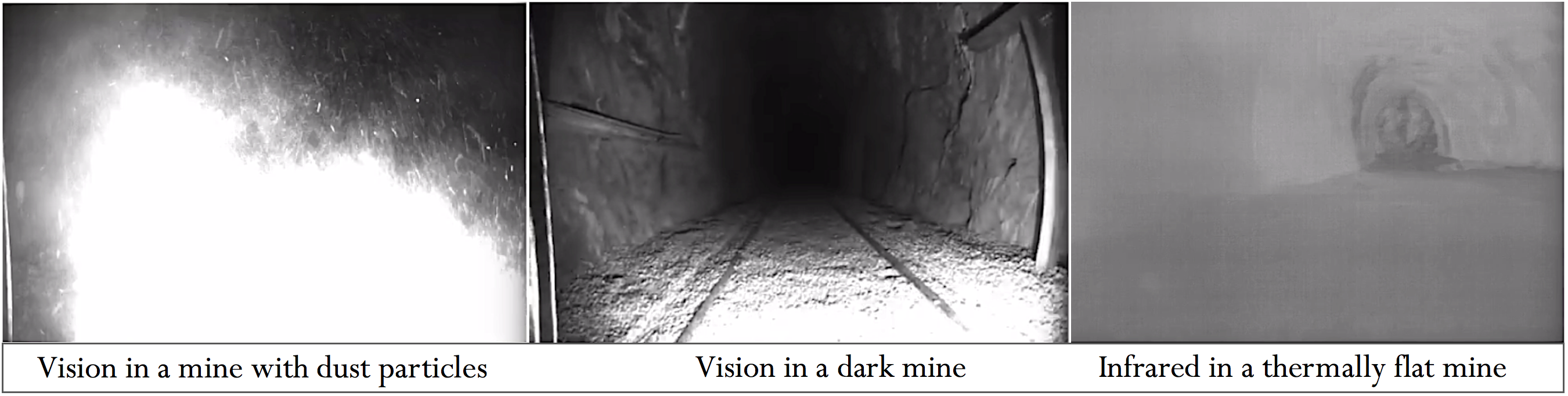}
	\caption{Examples of challenges introduced by perceptually-degraded subterranean environments to commonly used perception systems.}
	\label{fig:vision_underground} 
\end{figure}

Mobile robots rely on a model of the environment for navigation, collision avoidance and path planning. 
In GPS-denied and extreme unknown environments where no prior map of the environment is available, robots use the onboard sensing to construct locally accurate maps that can be used for navigation~\cite{SPOT}.
Multi-robot systems have been increasing in popularity over the past few decades~\cite{Choudhary, Carlone3} due to their potential advantages; higher performance and efficiency in performing spatially distributed tasks such as exploration and mapping a large-scale unknown environment, higher fault-tolerance and information redundancy, and scalability.
While great progress has been made over the past decades in the field of single-robot SLAM~\cite{Cadena}, extending these approaches to multi-robot systems in large-scale and ambiguous environments still remains a challenge.
As illustrated in Fig. \ref{fig:vision_underground}, perceptually-degraded environments are characterized by poor illumination and sparsity of salient perceptual and geometric features. Noisy measurements obtained in these environments lead to greater accumulation of errors in estimated robot trajectory which subsequently makes detection of intra- and inter-robot loop closures more challenging. This in turn affects the quality and consistency of constructed global maps. Robust and reliable localization and mapping in these environments can enable a wide range of terrestrial and planetary applications~\cite{AGUFall2019}, ranging from disaster relief in hostile environments to robotic exploration of lunar and Martian caves that are of particular interest as they can provide potential habitats for future manned space missions~\cite{Haruyama}.

SLAM algorithms aim to recover the most probable robot trajectory and map of the environment using the onboard sensing system.
Many solutions have been proposed based on different sensing modalities, including vision~\cite{ORB-SLAM}, visual-inertial~\cite{LOAM, Bloesch, Leutenegger}, thermal-inertial~\cite{Khattak1}, lidar-inertial~\cite{LION}, range-based SLAM~\cite{UWB}, or combination of them~\cite{LOCUS, HERO}. 
Vision-based localization and place recognition is a popular and well studied problem in the robotics and computer vision literature~\cite{vision-based-navigation-survey}, and while great progress has been made in visual loop closure detection~\cite{ORBSLAM, Lopez}, vision-based methods are faced with many challenges in perceptually-degraded environments including reduced visibility (i.e., inconsistent illumination or lack-thereof, fog, and dust), and change in the appearance, caused by viewing angle variations between different visits. Moreover, sparsity of salient perceptual features can lead to data association ambiguity and perceptual aliasing~\cite{Kamak1,Kamak2}.
Cameras and lidars are often used in conjunction due to their complementary nature~\cite{vision-lidar-lc, vision-lidar-lc2}. A 3D Lidar is a popular solution to mapping perceptually-degraded environments, as it does not rely on external light sources, and can provide range data over a $360^\circ$ horizontal field of view at a high temporal and spatial sampling rate.

Many scan registration methods have been developed for pose estimation from lidar data. Since raw lidar scans typically contain a large number of noisy and redundant points, methods are needed to downsample the point clouds in order to achieve real-time odometry and mapping performance with higher accuracy.
Nuchter et al.~\cite{Nuchter} present a lidar-based 6D SLAM algorithm that is used on a mine inspection robot. In order to achieve real-time performance, the authors propose a fast filtering method based on combining a median and a reduction filter to achieve significant data reduction in lidar scans while maintaining the surface structure in subterranean environments. 
Zhang et al~\cite{LOAM, LOAM2} present lidar odometry and mapping (LOAM) that relies on feature point extraction and matching to achieve real-time performance while minimizing motion distortions. 
While LOAM achieves good performance, it currently does not recognize loop closures which could help reduce the drift in estimated robot trajectory.
Ji et al.~\cite{LLOAM} present a 3D lidar mapping algorithm similar to LOAM that relies on segment based mapping and place recognition~\cite{SegMatch, SegMap} in 3D point clouds to detect loop closures in structured street environments.

Shan et al.~\cite{LeGoLOAM} propose the lightweight and ground-optimized (LeGO-LOAM) algorithm, an extension to LOAM that relies on the ground segmentation capability to discard points that may represent unreliable features. 
While their proposed method maintains the local consistency of the ground plane between consecutive frames, it does not use a global ground constraint that makes it susceptible to the accumulation of rotational error. In large-scale underground environments with uneven terrain, this could lead to large errors in the estimated robot poses and trajectories.
In LeGO-LOAM the search for loop closures is performed locally based on the estimated robot poses. This could lead to missed loop closure opportunities when drift in the robot trajectory is significant.
Hess et al.~\cite{Cartographer}, present Cartographer with real-time 2D mapping and loop closure capability. By combining a few consecutive lidar scans into a local submap, loop closures are detected by matching a large set of submaps. Upon detection of loop closure candidates, outlier loop closures are rejected by relying on the Huber loss in Sparse Pose Adjustment~\cite{SPA}.

Over the past decades a number of efforts have focused specifically on localization and mapping in underground mines and tunnels.
Thrun et al. \cite{Thrun} propose a SLAM algorithm for volumetric mapping of large underground mines. In order to reduce the drift in robot trajectories, the method uses a modified version of the Iterative Closest Point (ICP) \cite{ICP} algorithm to detect loop closures.
Tardioli et al.~\cite{Tardioli2} propose a system for exploration of a tunnel using a robot team. The system relies on a feature-based robot LOcalization Module (LOM) that is responsible for global localization by relying on pre-existing two-dimensional laser-segmentation-based maps. Global localization is achieved by matching observations of the environment with the map features.
In a similar work, Tardioli et al.~\cite{Tardioli3} propose an underground localization algorithm to enable autonomous navigation of a commercial dumper commonly used for underground construction. Localization is achieved by relying on semantic feature recognition in the tunnels to provide local information sufficient for a successful localization.

Zlot et al.~\cite{Zlot} present a 3D SLAM solution consisting of a spinning 2D lidar and an industrial-grade MEMS IMU to map a $17$ km long underground mine. In order to reduce the drift in open-loop trajectory, the authors rely on a global optimization algorithm where the entire robot trajectory is optimized given the set of computed constraints. Furthermore, a set of anchor points that are rough locations manually extracted from the 2D mine survey are used to further reduce the drift.
Leingartner et al.~\cite{Leingartner} evaluate how well off-the-shelf sensors and mapping solutions work in two different field experiments, one from a disaster-relief operation in a $1.5$ km long motorway tunnel, and one from a mapping experiment in a partly closed down tunnel. The authors conclude that despite advances in visual SLAM systems, lidar-based solutions have a superior localization and mapping performance in dark and perceptually-degraded underground tunnels.
Jacobson et al.~\cite{Jacobson} propose a monocular SLAM system for online localization and mapping in an underground mine with minimal human intervention. Their method leverages surveyed mine maps to increase mapping accuracy and enable global localization. In order to reduce the drift in the estimated robot trajectory, a vision-based loop closure detection method is used where a new frame is compared to all previously obtained frames for place recognition.

The earliest laser-based loop closure detection methods were based on registration of shape of the laser scans~\cite{Cox, Gutmann, FLU2}.
Cox et al.~\cite{Cox} proposed a method designed for use in structured office and factory environments where points were aligned with line segments extracted from a known environment representation. Gutmann et al.\cite{Gutmann} developed a scan-to-scan alignment approach that relied on the extraction of line segments from reference scans. 
Lu et al.~\cite{FLU2} improved the point-to-line matching method of Cox with an ICP-based point-to-point matching approach that do not require feature extraction or segmentation.
Moreover, similar to feature-based registration of 2D images~\cite{Harris}, several methods based on feature extraction from lidar point cloud have been proposed~\cite{LOAM, fpfh, LiOlson, Tipaldi} where the scan registration is achieved through matching the computed features descriptors.

Over the past decades the ICP algorithm has been commonly used for lidar scan registration and loop closure detection~\cite{LeGoLOAM, Bosse, Dorit, Bing, LAMP}.
In order to detect a loop closure, a lidar scan is registered to previously obtained scans to find a match. In large-scale or long-term operations where a large number of lidar scans are obtained over time, not only this method can become increasingly computationally expensive, but it can also lead to spurious or inaccurate loop closures, particularly in ambiguous environments.
In order to improve the performance and accuracy of loop closures, a common approach is to constrain the search for loop closures to a fixed radius centered at estimated robot poses. While this method is effective in reducing the computational load and number of spurious loop closures, it could lead to missed loop closure opportunities when the accumulation of errors in lidar odometry leads to significant drift in the robot trajectory.

This paper extends our previous work on large-scale autonomous mapping and positioning (LAMP)~\cite{LAMP} in a number of significant ways to improve localization and mapping in perceptually-degraded environments. The core contributions of this paper are:
\begin{enumerate}
    \item A real-time degeneracy-aware SLAM front-end to determine the level of geometric degeneracy in unknown environments. Using this crucial capability, ambiguous areas that could lead to data association ambiguity and spurious loop closures are identified and excluded from the search for loop closures. This significantly improves the quality and accuracy of loop closures because the search for loop closures can be expanded as needed to account for drift in robot trajectory, while decreasing rather than increasing the probability of spurious loop closures.
    \item A drift-resilient and pose-invariant loop closing method based on salient 2D and 3D geometric features extracted from lidar point cloud data to enable detection of loop closures with increased robustness and accuracy as compared to traditional geometric methods. The proposed method does not require additional sensors to achieve higher performance, and instead relies on exploiting salient features embedded in 2D occupancy grid maps commonly used in robot navigation to obtain more information about the spatial configuration of the local environment from lidar point cloud data. The method is not dependent on the estimated robot poses, a key advantage that makes it invariant to drift in robot trajectories. In the first stage of the process, 2D features extracted from salient occupancy grid maps are used to search for potential loop closures over the entire robot trajectory. This significantly reduces the number of missed loop closure opportunities as compared to geometric methods that perform the search for loop closures locally. In the second stage, a geometric verification step is used to verify the quality of each loop closure candidate. Subsequently, a back-end equipped with an outlier rejection capability performs the map merging and alignment.
    \item An extensive evaluation and comparison of performance with state-of-the-art methods is provided in a variety of extreme subterranean environments. This includes data collected at the Tunnel Circuit of the DARPA Subterranean Challenge\cite{SubT} in the Bruceton Safety Research coal mine and Experimental mine in Pittsburgh, PA.
\end{enumerate}

The rest of this paper is organized as follows: in Section \ref{sec:problem_formulation} we present the problem formulation. Our lidar-based SLAM architecture which includes the degeneracy-aware SLAM front-end, the drift-resilient loop closing pipeline, and the back-end is presented in Section \ref{sec:methodology}. Experimental results are presented in Section \ref{sec:Experiments}. Finally, Section \ref{sec:discussion} discusses the conclusion, and future research directions.

\section{Problem Formulation} \label{sec:problem_formulation}
In extreme and unknown environments with sparse salient geometric structures, the lidar odometry is challenged with a higher probability of data association ambiguity that could lead to large errors when estimating the relative robot motion. Robust loop closure detection is crucial to improve localization and mapping accuracy in these environments. Moreover, in a multi-robot collaborative SLAM system, loop closures are critical to find correspondences between the maps obtained by individual robots in order to merge them into a consistent global map of the environment.

\begin{figure}[t!]
\centering 
  \includegraphics[width=1\columnwidth]{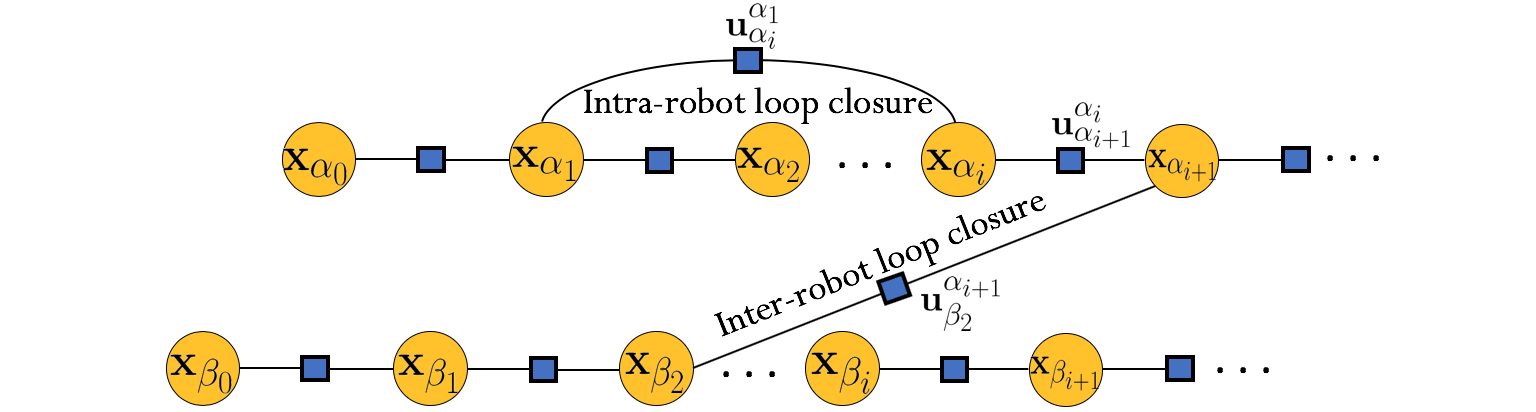}
  \caption{Illustration of multi-robot pose graph SLAM for a team of two robots. The yellow nodes correspond to the robot poses, while blue squares represent the relative 3D rigid transformation between the nodes. \label{fig:simple_posegraph}}
\end{figure}

The focus of this work is developing methods for improving location ambiguities and loop closing in extreme environments that do not provide sufficient distinctive information to establish global position.
While our proposed method can be valuable to map alignment and merging in multi-robot systems, it naturally falls back to loop closing for robust localization and mapping in a single-robot system.
As illustrated in Fig. \ref{fig:simple_posegraph}, in order to represent a robot trajectory and the map of the environment, a graph-based formulation is used~\cite{Cadena}, where every node in the graph corresponds to a robot pose, and every edge connecting two nodes expresses the relative 3D motion between the corresponding poses. 
We denote the trajectory of $N_{\alpha}$ poses of robot $\alpha$ by the sequence $\textbf{x}_{\alpha} \doteq [\textbf{x}_{\alpha_i}]_{i = 1:N_\alpha}$, where $\textbf{x}_{\alpha_i} \doteq [\textbf{R}_{\alpha_i}, \textbf{t}_{\alpha_i}] \in$ SE(3) is the robot pose associated with the $i$-th node in the graph, which includes a rotation $\textbf{R}_{\alpha_i} \in$ SO(3), and a translation $\textbf{t}_{\alpha_i} \in \mathbb{R}^3$.

The lidar point cloud obtained at each pose is associated with the corresponding node in the pose graph.
Each robot builds an explicit map of the environment by projecting the point clouds associated with the nodes into a common world coordinate system $\mathcal{W}$.
Considering a centralized multi-robot system, when a robot is within the wireless communication range, it communicates its local pose graph to a base station. Assuming a partial overlap among areas explored by all robots, the base station is responsible for merging the local graphs to produce a consistent global map of the environment that can be used for reliable navigation of robots in the future.

The set of all raw lidar scans obtained along the trajectory of robot $\alpha$ is denoted by $z_{\alpha} = \{z_{\alpha_i}\}_{i=1:N_\alpha}$, where $z_{\alpha_i} = \{p_k\}_{k=1:N_z}$ is the $i$-th lidar scan represented as a collection of 3D points $p_k \in \mathbb{R}^3$ in the local lidar coordinate system $\mathcal{L}_i$. We denote the measurement model by
\begin{align} \label{Eq:observation_noise}
    z_{\alpha_i} = h(\textbf{x}_{\alpha_i}, m_{\alpha}) + w_{\alpha_i},
\end{align}
where $h(.)$ returns the vector connecting the current lidar pose to a point in the environment along each lidar ray, $m_\alpha$ represents the actual local environment explored by robot $\alpha$, and $w_{\alpha_i} \sim \mathcal{N}(0, \Sigma_{\alpha})$ is assumed to be a Gaussian noise with zero-mean and covariance $\Sigma_{\alpha}$.

In lidar-based state estimation, the odometric estimates are obtained by computing the relative pose transformation between consecutive lidar scans. 
In this paper, we use the Generalized ICP algorithm~\cite{GICP} to obtain $\textbf{u}^{t}_{t+1} \doteq [\textbf{R}^{t}_{t+1}, \textbf{t}^{t}_{t+1}]$, an odometric estimate that encodes the relative 3D motion between two consecutive lidar scans obtained at times $t$ and $t+1$.
As the lidar operates at a high temporal sampling rate relative to robot motion, the pose graph would be oversampled in time if a new node were instantiated after each new odometric measurement.
We utilize a reduced pose graph where a new node is instantiated in the graph after reaching a minimum odometric displacement ($30^\circ$ rotation or $1$ m translation). We refer to the nodes in the reduced graph as \emph{key-nodes}, and the lidar scans associated with them as \emph{key-scans}. The relative 3D motion $\textbf{u}^{i}_{{i+1}}$ between poses associated with key-nodes $i$ and ${i+1}$ is obtained from integration of incremental motion estimates between times $t$ and $t+n_i$.

Since the lidar measurements are noisy, each lidar-based odometric estimate will include some error as the result of point cloud matching errors. Since the robot trajectory is built incrementally from the accumulation of odometric estimates, the accumulation of translational and angular errors in a sequence of odometric estimates can lead to an unbounded drift in the estimated robot trajectory for an assumed unlimited time. 
In order to bound and reduce this drift, our method relies on a degeneracy-aware and saliency-based loop closing method, where we constrain the search for loop closures to only the most observable areas in the unknown environment. 
The saliency-based loop closing pipeline can be abstractly formulated as
\begin{equation} \label{Eq:two_stage}
LC^g(LC^s(\alpha_i,\beta_j)) =
\begin{cases}
  0 & \ \text{if} \ \ \text{no loop closed} \\
  1 & \ \text{if} \ \ \text{loop closed}
\end{cases},
\end{equation}
where $LC^s(.,.)$ and $LC^g(.,.)$ denote the stages of \emph{pre-matching} and \emph{geometric verification} based on salient 2D and 3D features extracted from lidar point cloud data respectively. 

In the first stage, 2D occupancy grid maps are used to create bird's eye views of point cloud data to extract additional information about the spatial configuration of local scenes. This enables the pre-matching stage $LC^s(.,.)$, where putative loop closures are identified by evaluating the similarity between spatial configuration of fully observable scenes along the robot trajectory.
Upon detection of putative loop closures, the geometric verification stage $LC^g(.,.)$ is used to evaluate the quality of each loop closure candidate, and to compute the 3D rigid transformation between the pair of lidar scans. When $LC^g(LC^s(\alpha_i,\beta_j)) = 1$, an edge $\textbf{u}^{\alpha_i}_{\beta_j}$ connecting the nodes $x_{\alpha_i}$ and $x_{\beta_j}$ is added to the pose graph to represent a loop closure.
When $\alpha = \beta$, it represents an intra-robot loop closure found in a local pose graph, otherwise, it represents an inter-robot loop closure found between two local pose graphs.

In a single-robot SLAM system, the goal is to estimate the unknown robot trajectory and the map of the environment from the set of all odometry and loop closure estimates. 
Let $G = \langle \textbf{x}, \textbf{u}, \textbf{z} \rangle$ denote the local pose graph comprised of the robot trajectory $\textbf{x}$, the set of all odometric and intra-robot loop closure estimates $\textbf{u}$, and the set of all key-scans $\textbf{z}$. Since the process happens on the same robot, the robot name is dropped for brevity.
The maximum a-posteriori (MAP) estimate of the robot trajectory can be obtained using Pose Graph Optimization (PGO)~\cite{Cadena} as given by
\begin{align} \label{Eq: PGO}
\centering
    \hat{\textbf{x}} & = \argmax_{\textbf{x}} p(\textbf{x} | \textbf{u}) \nonumber \\ 
    & = \argmax_{\textbf{x}} p(\textbf{x})p(\textbf{u} | \textbf{x}),
\end{align}
where $p(\textbf{x})$ is the prior probability over $\textbf{x}$, and $p(\textbf{u}| \textbf{x})$ is the measurement likelihood.
When no prior knowledge is available over the robot trajectory, the term $p(\textbf{x})$ is assumed to be a uniform distribution and does not affect the result. This will reduce the MAP estimation problem to the maximum likelihood (ML) estimation as given by
\begin{align} \label{Eq: ML_estimation}
\centering
    \hat{\textbf{x}} & = \argmax_{\textbf{x}}\prod_{i=1}^{|\textbf{u}|}p(\textbf{u}_{i} | \textbf{x}_{i}) \nonumber \\
    & = \argmin_{\textbf{x}} -log \Big(\prod_{i=1}^{|\textbf{u}|}p(\textbf{u}_{i} | \textbf{x}_{i})\Big) \nonumber \\
    & = \argmin_{\textbf{x}} -log \Big(\prod_{i=1}^{N} \underbrace{p(\textbf{u}^{i}_{i+1} | \textbf{x}_{i}, \textbf{x}_{{i+1}})}_\text{Odometry} \prod_{\textbf{u}^{i}_{j} \in \textbf{u}_{c}} \underbrace{p(\textbf{u}^{i}_{j} | \textbf{x}_{i}, \textbf{x}_{j})\Big)}_\text{Loop Closure},
\end{align}
where $N$ is the total number of nodes in the graph, and $\textbf{u}_{c}$ denotes the set of all loop closure constraints.
Let $\textbf{x}_{j} = f(\textbf{x}_{i},\textbf{u}^{i}_{j}) + v_{i,j}$ be the new pose update using the robot motion model from node $i$ to $j$,
where $f(.)$ computes robot's nonlinear motion between the unknown robot poses $x_{i}$ and $x_{j}$, and $v_{i,j}$ is assumed to be a zero-mean Gaussian noise with information matrix $\Omega_{i,j}$. The ML trajectory estimate can be computed by minimizing the mismatch between the relative pose measurements and the estimated robot poses as given by
\begin{equation}
\begin{split} 
    \hat{\textbf{x}} = \argmin_{\textbf{x}} \big(\sum_{i = 1}^{N}\norm{f(\textbf{x}_{i},\textbf{u}^{i}_{i+1}) - \textbf{x}_{i+1}}^2_{\Omega_{i,i+1}} + 
    \sum_{\textbf{u}^{i}_{j} \in \textbf{u}_{c}|}\norm{f(\textbf{x}_{i},\textbf{u}^{i}_{j}) - \textbf{x}_{j}}^2_{\Omega_{i,j}}\big),
\end{split}
\end{equation}
where $\norm{.}^2_{\Omega_{i,j}}$ denotes the squared Mahalanobis distance.

To extend this to a centralized multi-robot system, assuming known initial robot poses, a similar process is used on the base station to detect inter-robot loop closures to determine correspondences between pose graphs.
Once the relative 3D rigid transformations between overlapping parts of robot trajectories are determined, pose graph optimization (\ref{Eq: PGO}) is used to obtain the maximum a-posteriori (MAP) estimate of the global pose graph. The global map of the environment can then be constructed by projecting the key-scans into the common world coordinate system $\mathcal{W}$.

\section{Methodology} \label{sec:methodology}
\subsection{Degeneracy-Aware SLAM Front-End} \label{sec:lidarOdometry} 
\begin{figure}[b!]
\centering
	\includegraphics[width=1.0\columnwidth]{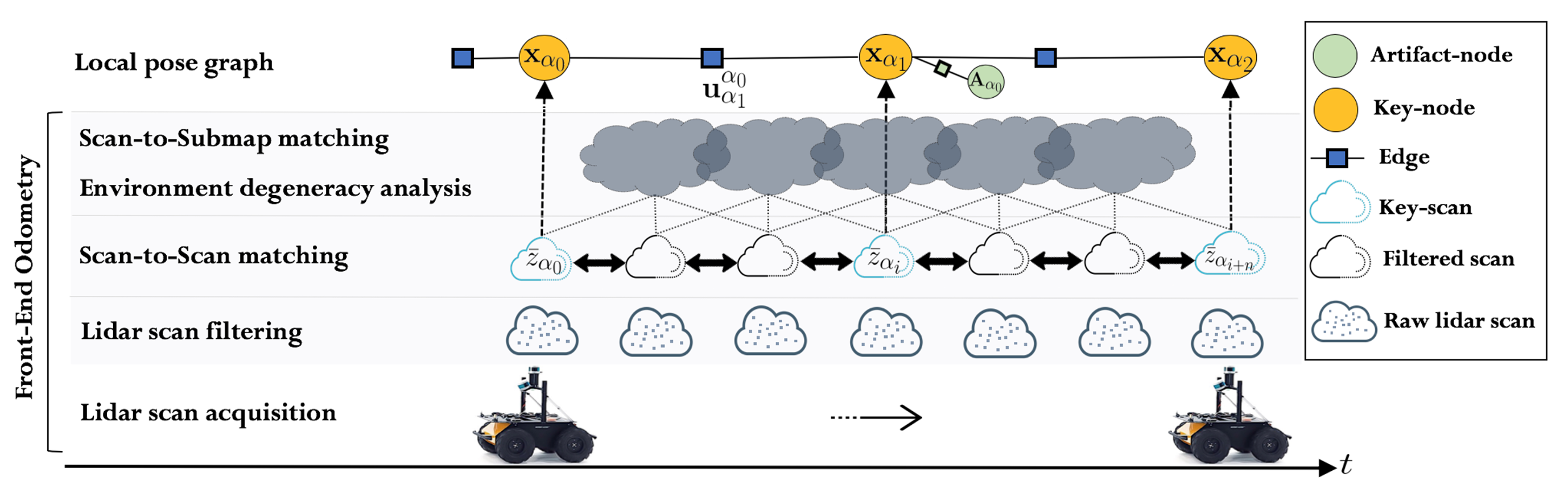}
	\caption{Overview of the lidar-based front-end and the local pose graph.}
	\label{fig:system_overview}
\end{figure}

The lidar-based SLAM front-end is responsible for data association from sensor measurements to produce odometric estimates and loop closure constraints.
The odometric estimates are obtained by computing the relative 3D motion that best aligns consecutive lidar scans. Fig.~\ref{fig:system_overview} provides an overview of the front-end that has three layers of processing to produce odometric estimates: (i) point cloud filtering, (ii) scan-to-scan registration, and (iii) scan-to-submap registration. In the following we provide a short description of each layer.

The raw point clouds obtained by each scan are often density-uneven and contain redundant and noisy points that can lead to increased computational load and inaccurate odometric estimates using the ICP algorithm.
To overcome these challenges, most methods\cite{Nuchter, LOAM,LLOAM,LeGoLOAM,LAMP} use a subset of the raw point cloud data to estimate the robot motion.
Among different point cloud filtering methods, random downsampling, and voxel grid filtering~\cite{LLOAM, Random_filter, pointCloud_filter} are commonly used for point cloud filtering. In random downsampling method, points are randomly sampled with a uniform probability, whereas, in voxel grid filtering, a 3D voxel grid is created over the point cloud data, and the points in each voxel are approximated with their centroid value. 
While these filtering approaches are effective in downsampling the raw point cloud data, they remove points regardless of their importance and effect on the performance and accuracy of lidar odometry.

In this work, using the method of Zhang et al.~\cite{LOAM}, we decimate up to $90\%$ of the points in every raw lidar scan by extracting a set of uniformly distributed salient features located on sharp edges and planar surfaces.
We then use the GICP algorithm, to obtain odometric estimates based on a two-stage scan-to-scan and scan-to-submap registration process as initially introduced in our prior work, LAMP \cite{LAMP}. In the scan-to-scan registration step, the incoming point cloud is aligned to the last obtained lidar scan to find a rigid body transformation that best aligns the two point clouds. 
After this approximate localization, the solution is further refined by registering the incoming point cloud to a submap created from the local region of previously collected point cloud data in the pose graph. Given the current estimated pose of the robot in the world coordinate system $\mathcal{W}$, the incoming lidar scan is transformed to the world coordinate system using the same transformation, and an approximate nearest neighbors search is performed to find the closest points in the map. A second point cloud registration similar to scan-to-scan registration is used to obtain the odometric estimate $\textbf{u}^t_{t+1} \doteq [\textbf{R}^t_{t+1}, \textbf{t}^t_{t+1}] \in SE(3)$ that describes the 3D motion of the robot between times $t$ and $t+1$.

\begin{figure}[b!]   
\centering 
	\includegraphics[width=1.0\columnwidth]{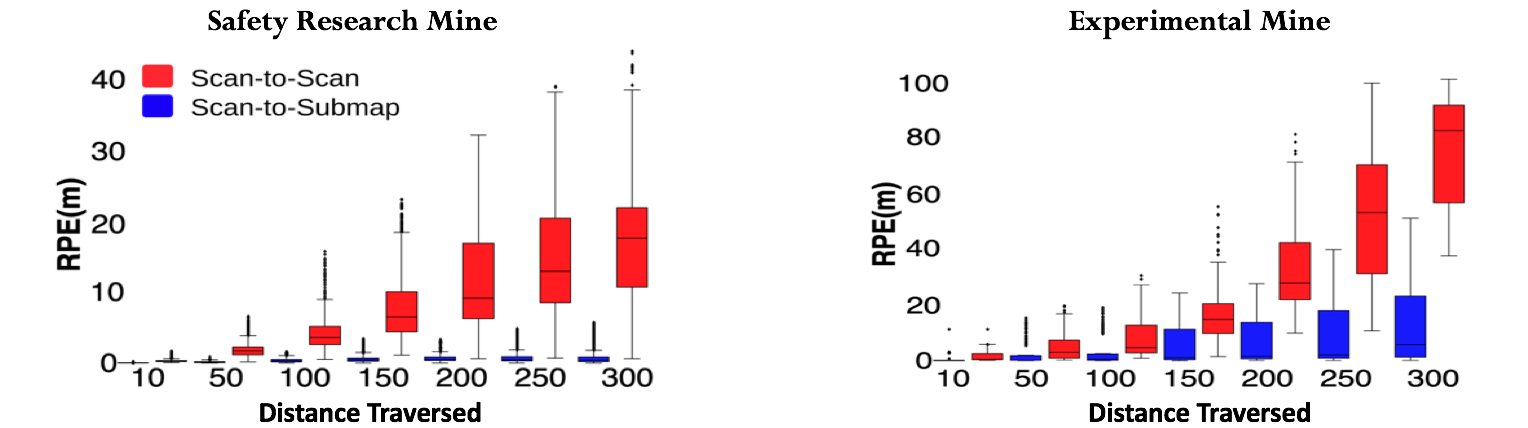}
	\caption{Quantitative evaluation of drift from scan-to-scan and scan-to-submap matching as a function of distance traversed in Bruceton Safety Research and Experimental mines. Each box comprises the RPE values ranging from the first to the third
    quartile. The median is indicated by the black horizontal bar. The whiskers extend to the farthest data points that are within $1.5$ times the interquartile range. Outliers are shown as dots.}
	\label{fig:frontend-drift-qual-eval}
\end{figure}
Fig.~\ref{fig:frontend-drift-qual-eval} reports the accuracy of lidar odometry by measuring the Relative Pose Error (RPE) as a function of distance traversed using the package for evaluation of odometry and SLAM (EVO)~\cite{Grupp} in two perceptually-degraded underground mines.
\begin{figure}[t!]   
\centering 
	\includegraphics[width=1.0\columnwidth]{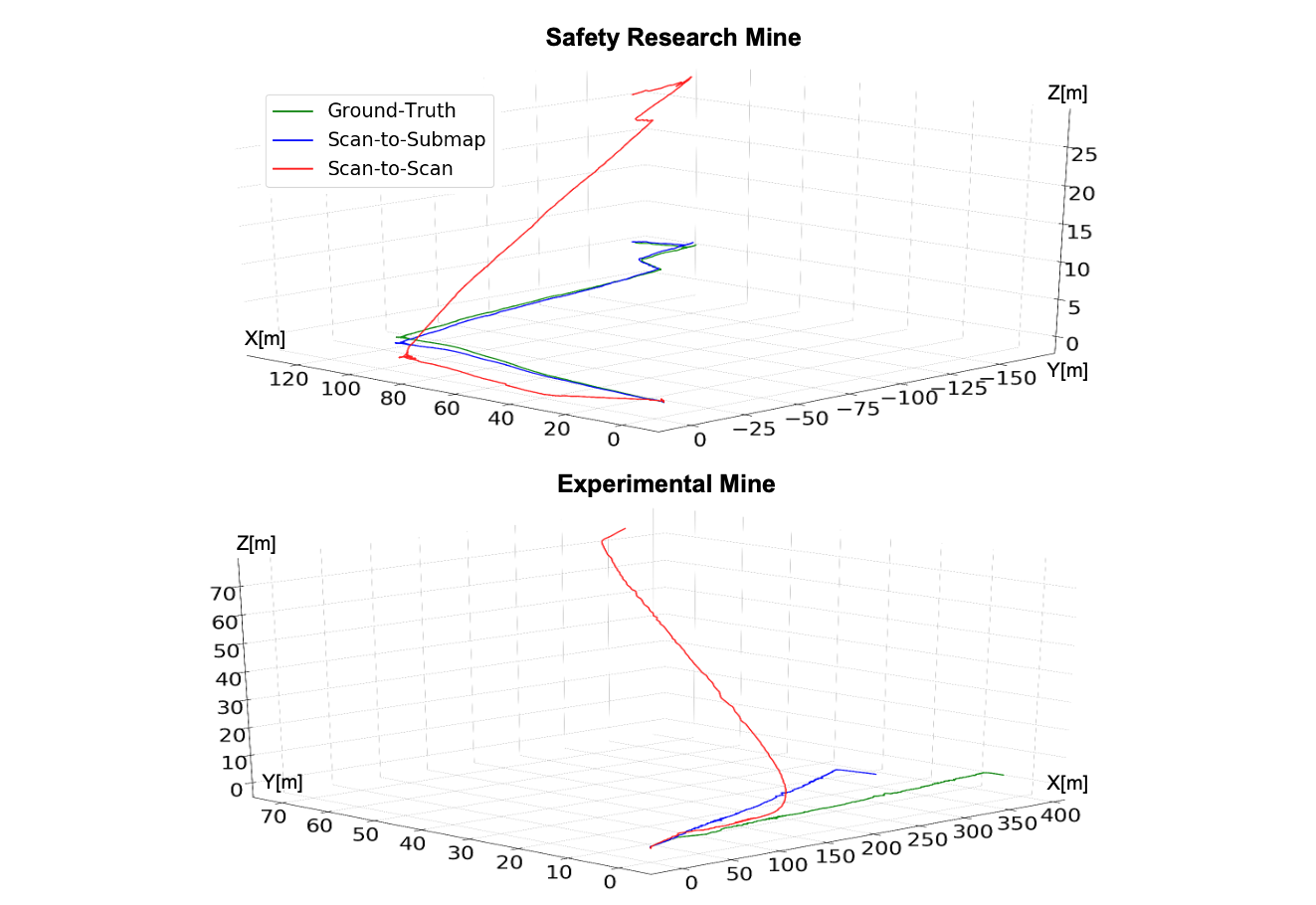}
	\caption{Robot trajectories obtained using scan-to-scan and scan-to-submap registration methods in Safety Research mine and Experimental Mine.}
	\label{fig:odom_eval_plots}
\end{figure}
In the Safety Research mine, the robot autonomously navigates $1400$m of the mine tunnels.
By relying on scan-to-scan registration, the accumulation of errors results in more than $18$m ($6\%$ relative position error) average drift in translation for each $300$m of traversed distance. By using the scan-to-submap registration step the RPE is reduced to less than $1\%$ of distance traversed for each $300$m of travel.

The Experimental mine presents a more challenging subterranean environment with long and featureless corridors. During the $700$m autonomous navigation, due to high level of geometric degeneracy at multiple locations, accumulation of errors from scan-to-scan registration leads to more than $80$m ($27\%$ relative position error) of average drift in translation for each $300$m of distance traversed. 
By using the scan-to-submap registration step, the estimated motion of the robot is further refined and the drift in translation is reduced to $7$m ($2\%$ relative position error) of average drift for each $300$m of traversed distance.
Fig. \ref{fig:odom_eval_plots} presents partial estimated robot trajectories in the Safety Research and Experimental mines using the scan-to-scan and scan-to-submap matching methods. The results show that due to low vertical resolution of the lidar scanner, scan-to-scan matching poorly estimates the angular and translational motion of the robot in the narrow tunnels of the mines. While this leads to significant drift in the estimated robot trajectories, the scan-to-submap matching step can reduce the drift by using an additional registration step where the output of the ICP algorithm is further refined. The Experimental mine represents a more challenging environment with many degenerate scenes due to long, narrow and symmetrical walls. The result show, while using the scan-to-submap matching step reduces the error in robot estimates, the drift is still substantially large. This highlights the importance of a drift-resilient loop closure detection method that can reduce the accumulated error in robot trajectory. 

\subsection{Determination of Geometric Degeneracy} \label{Sec:online_degen_eval}
Determination of reduced observability and geometric degeneracy is a crucial capability to evaluate the reliability of the front-end in reconstructing its full internal state based on measurements of the environment. 
Loss of observability can occur in any odometry system; scarcity of texture and salient features in a vision-based system, lack of thermal variations in a thermally flat environment (e.g., cold underground mines and caves) in thermal-inertial odometry, sparsity of distinctive geometric structures in a lidar-based front-end, or slippery terrains in a wheel-inertial odometry system. 
In lidar odometry, the reliability of odometric estimates is affected by the noise and degeneracy that can arise from the geometrical structure of the environment. For example, partial observability may occur when a robot traverses a tunnel or corridor with long, symmetrical and featureless walls. 
In this environment, the scan matching optimization process may result in multiple possible solutions leading to inaccurate estimation of forward motion along the direction of the tunnel.

In order to evaluate the level of geometric degeneracy in an unknown environment, we develop a method using a distance cost function based on the rigid transformation obtained from the ICP algorithm. 
From the last iteration of the ICP algorithm, we obtain the set of all corresponding points $P^t$ and $P^{t+1}$, and the relative pose measurement $\textbf{u}^{t}_{t+1} \doteq [\textbf{R},\textbf{t}]$ between consecutive lidar scans, where $\textbf{R}(\theta_x,\theta_y,\theta_z)$ is the 3D rotation defined in terms of Euler angles roll $\theta_x$, pitch $\theta_y$, and yaw $\theta_z$, and $\textbf{t} = [t_x,t_y,t_z]$ is the 3D translation vector. 

Given $\textbf{u}^t_{t+1}$ that best aligns two consecutive lidar scans, the alignment error between the $k$th pair of corresponding points $p^t_k$ and $p^{t+1}_k$ is given by
\begin{align} \label{Eq:misalignment_vector}
    \textbf{d}_k = \textbf{u}^t_{t+1}p^t_k - p^{t+1}_k,
\end{align}
where $\textbf{d}_k = [d_{k_x}, d_{k_y}, d_{k_z}]$ is the misalignment vector. 
The mean squared distance between the set of all corresponding points $P^{t}$ and $P^{t+1}$ is given by
\begin{align} \label{Eq:MSE}
    {\mathcal{E}}(\textbf{u}^t_{t+1}P^{t}, P^{t+1}) = \frac{1}{N_k}\sum_{k=1}^{N_k}\norm{d_k}^2,
\end{align}
where $\mathbf{\mathcal{E}}(.)$ is a scalar-valued cost function, $N_k$ is the total number of corresponding points from the ICP algorithm, and $\norm{.}$ denotes the $\mathcal{L}_2$ norm.
The ICP-based scan matching can be formulated as an optimization problem that aims to find $\textbf{u}^t_{t+1}$ that minimizes the cost function as given by
\begin{align} \label{Eq:scan-to-scan}
    \textbf{u}^t_{t+1} = \argmin_{\hat{\textbf{u}}^t_{t+1}}{\mathcal{E}}(\hat{\textbf{u}}^t_{t+1}P^{t}, P^{t+1}).
\end{align}

Using the displacement vector computed for the $k$-th pair of corresponding points in (\ref{Eq:misalignment_vector}), a cost function can be defined in terms of $\textbf{d}_k$,  as given by
\begin{align} \label{Eq:objective_function}
    \xi_k(\textbf{u}^{t}_{t+1}) = d^2_{k_x} + d^2_{k_y} + d^2_{k_z}.
\end{align}
Assuming local linearity due to small motion between consecutive lidar scans, we compute the gradient of the cost function in order to propagate errors from sensor noise to uncertainty of state estimates as given by
\begin{align}
J^{t+1}_k &= \frac{\partial \xi_k(\textbf{u}^{t}_{t+1})}{\partial (t_x, t_y, t_z, \theta_x, \theta_y, \theta_z)} \\ \vspace{4pt}
\notag
&= 
   \begin{bmatrix}
    \frac{\partial \xi_{k}(\textbf{u}^{t}_{t+1})}{\partial t_x} & \frac{\partial \xi_{k}(\textbf{u}^{t}_{t+1})}{\partial t_y} & \frac{\partial \xi_{k}(\textbf{u}^{t}_{t+1})}{\partial t_z} & \frac{\partial \xi_{k}(\textbf{u}^{t}_{t+1})}{\partial \theta_x} & \frac{\partial \xi_{k}(\textbf{u}^{t}_{t+1})}{\partial \theta_y} & \frac{\partial \xi_{k}(\textbf{u}^{t}_{t+1})}{\partial \theta_z}
  \end{bmatrix},
\end{align}
\normalsize
where $J^{t+1}_k$ is the gradient vector computed for the $k$th pair of corresponding points at time step $t+1$.
The Hessian of the cost function can be approximated~\cite{Szeliski} from the sum of outer products of gradients for the set of all $N_k$ corresponding points,
\begin{align} 
    H^{t+1} = \sum_{k=1}^{N_k} (J^{t+1}_k)^TJ^{t+1}_k.
\end{align}
In the eigenvalue decomposition of $H^{t+1}$, the eigenvector associated with the smallest eigenvalue represents the least observable direction in terms of the estimated rigid 3D transformation.
We find the relative scale between the most and least observable directions by computing the ratio of the maximal and minimal eigenvalues
\begin{align} \label{Eq:condition_number}
    \mathcal{\kappa}^{t+1} = \frac{\lambda_{max}}{\lambda_{min}},
\end{align}
where $\kappa^{t+1}$ is the condition number of the approximated Hessian $H^{t+1}$ which is a symmetric matrix.
Although the eigenvalues will be dependent on general environmental variables such as measurement noise or the number of corresponding points in the obtained lidar scans, computing the ratio will remove dependence on environmental variables that scale the eigenvalues. Thus a large condition number corresponds to higher levels of geometric degeneracy~\cite{condition_number}. This means a relatively large error is associated with one or more linear combination of parameters of the computed transformation $\textbf{u}^{t}_{t+1}$.
A small condition number close to $1$ corresponds to equal observability of all parameters in the computed transformation.
In our experiments, the pose is classified as degenerate if $\text{log}(\kappa^{t+1}) \geq \kappa_{th}$, where $\kappa_{th}$ is the degeneracy threshold based on expectations of normal variability, and validated by experimental results. 
Robot poses and corresponding key-scans that are classified as degenerate are removed from loop closure consideration to achieve the desired performance objective by constraining the search for loop closures to areas with highest level of observability to reduce the probability of false or inaccurate loop closures.

\begin{figure}[b!] \centering 
	\includegraphics[width=1.0\columnwidth]{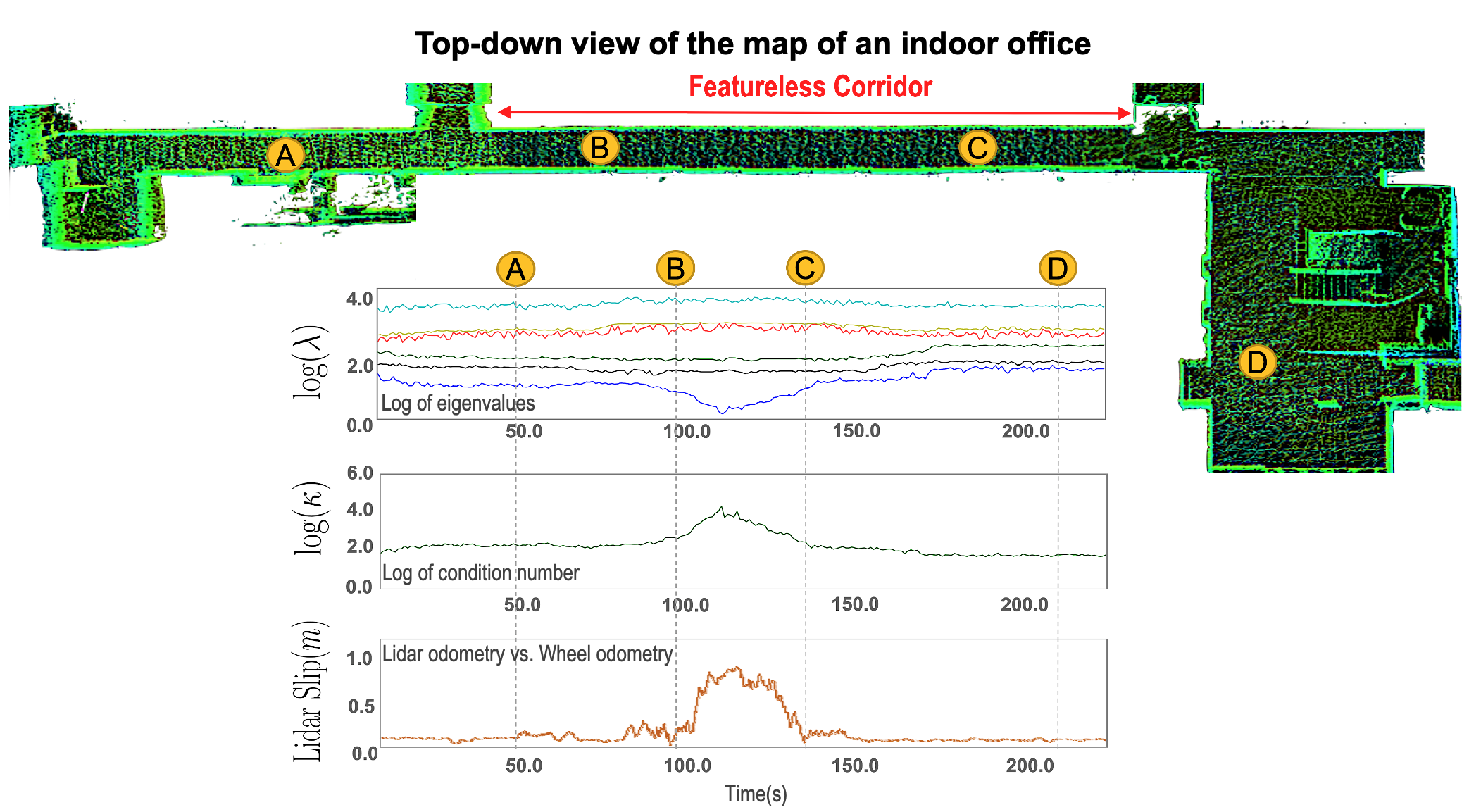}
	\caption{(a) map of a corridor in an indoor office. Several locations are marked with A, B, C, and D to show the response of eigenvalues and condition number to different levels of geometric degeneracy in the scene. $\log{\lambda}$ shows the response of all six eigenvalues of the approximated Hessian as the robot navigates the corridor. $log({\kappa})$ plot shows the log of condition number computed from ratio of the maximal and minimal eigenvalues. The lidar slip plot shows the difference between forward translation of the robot as estimated by the wheel-inertial odometry and lidar odometry at $1$s intervals. The wheel slippage in this experiment is known to be negligible.}
	\label{fig:degenerecy}
\end{figure}
Fig. \ref{fig:degenerecy} shows the determination of geometric degeneracy in an indoor office environment as a robot navigates a long corridor with flat and symmetric walls.
The plot of log of the eigenvalues $log(\lambda)$ shows the response of all six eigenvalues of the approximated Hessian matrix. 
As the robot navigates the featureless corridor, in one section of the trajectory a significant drop can be seen in the values of the smallest eigenvalue of the Hessian, while variations in the rest of eigenvalues are minimal. This leads to an increase in the values of the condition number in (\ref{Eq:condition_number}) as shown in the plot of $\text{log}(\kappa)$ values.
In this experiment, it was possible to verify that increased values of $\text{log}(\kappa)$ corresponded to noisy lidar-based odometric estimates based on the wheel-inertial odometry. 
We compare the output of the lidar odometry with the wheel-inertial odometry as the robot navigates the indoor office environment. Often the wheel odometry is not reliable, but as the robot moves at a low speed and the corridor has carpet flooring, the wheel slippage is known to be negligible. 
The lidar slip plot shows the difference in the computed forward motion based on the wheel-inertial and lidar odometry at $1$s intervals. 
The results show values of $\text{log}(\kappa) > 2.0$ correspond to the largest discrepancies between the lidar and wheel-inertial odometry. The degeneracy metric $\kappa$ responds to the geometric degeneracy of the featureless corridor, as the forward motion of the robot along the walls becomes unobservable to the lidar odometry: a situation we call \textit{lidar-slip}.

\begin{figure}[h!]   
\centering 
	\includegraphics[width=1\columnwidth]{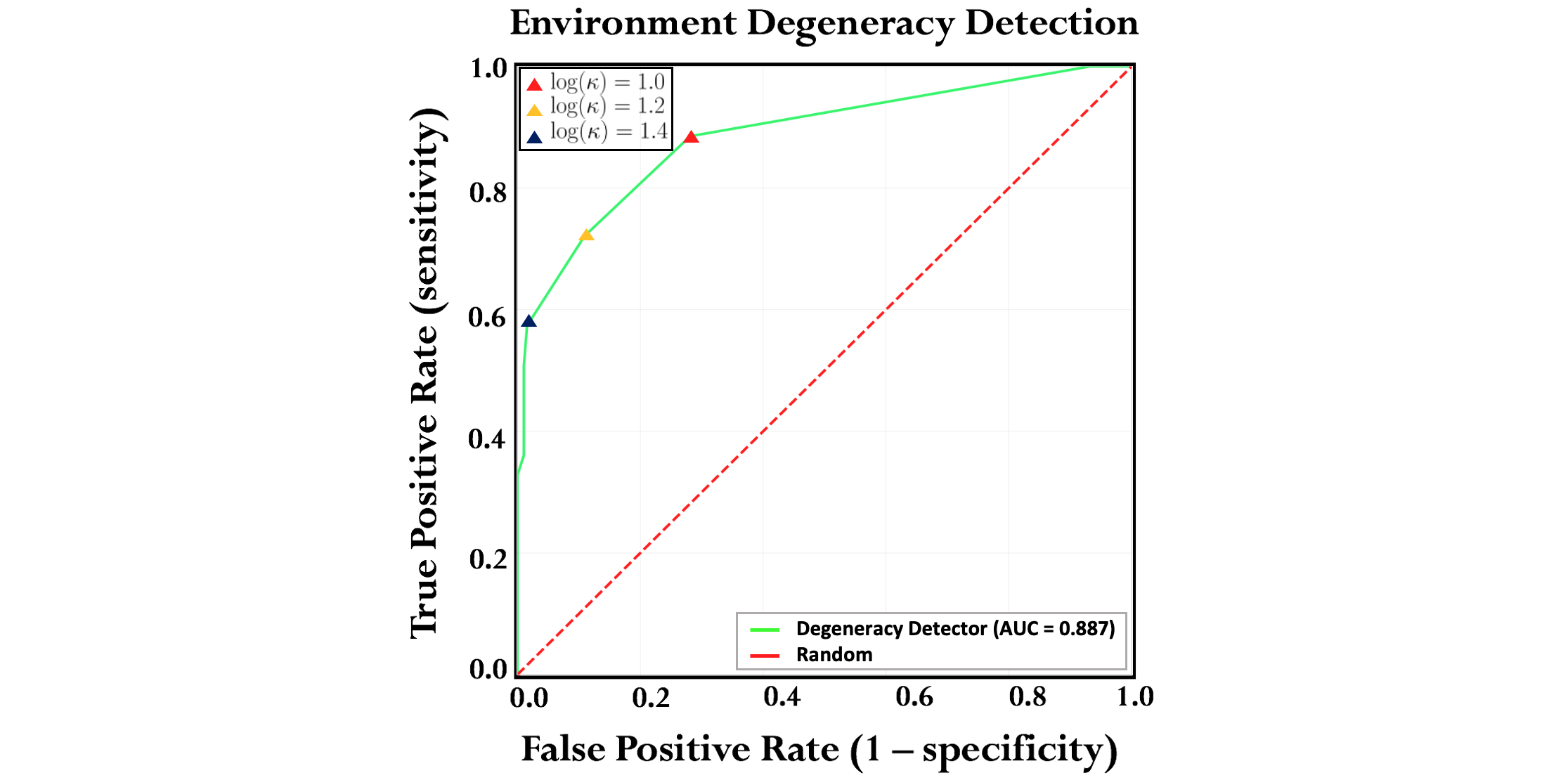}
	\caption{ROC plot reports the performance of environment degeneracy detector, created by plotting the true positive rate (TPR) against the false positive rate (FPR) at various $log(\kappa)$ threshold settings.} 
	\label{fig:roc}
\end{figure}

In order to evaluate the sensitivity and specificity of geometric degeneracy detector, we rely on Receiver Operating Characteristic (ROC) analysis. A set of $254$ lidar scans are manually labeled such that $61$ lidar scans correspond to degenerate scenes. 
The ROC curve is produced by computing the true positive and false positive rates at various thresholds. 
Fig. \ref{fig:roc} reports the high discriminative ability of geometric degeneracy detector where an area under curve (AUC) value of $0.887$ is achieved. Different decision thresholds marked on the plot show the sensitivity of the method to different thresholds. 

\subsection{Loop Closure Detection}\label{sec:lidarLC} 
In this section, we first study the shortcomings and challenges of ICP-based loop closures that are commonly used in some of the state-of-the-art methods~\cite{LeGoLOAM, Bosse, Dorit, Bing, LAMP}, and then propose a degeneracy-aware and saliency-based method to improve place recognition and loop closing in large-scale and perceptually-degraded environments.

In the absence of any prior map of the environment that can be used for global localization, the open loop accumulation of errors from lidar odometry can lead to an unbounded drift in the estimated robot trajectory as error variance increases after each motion step. This drift, is inherent to any open loop odometry system, and it can only be limited and subsequently reduced by detection of loop closures when a robot returns to a previously visited or otherwise known location or landmark.
\begin{figure}[b!]
\centering
	\includegraphics[width=1.0\columnwidth, trim= 0mm 0mm 0mm 0mm, clip]{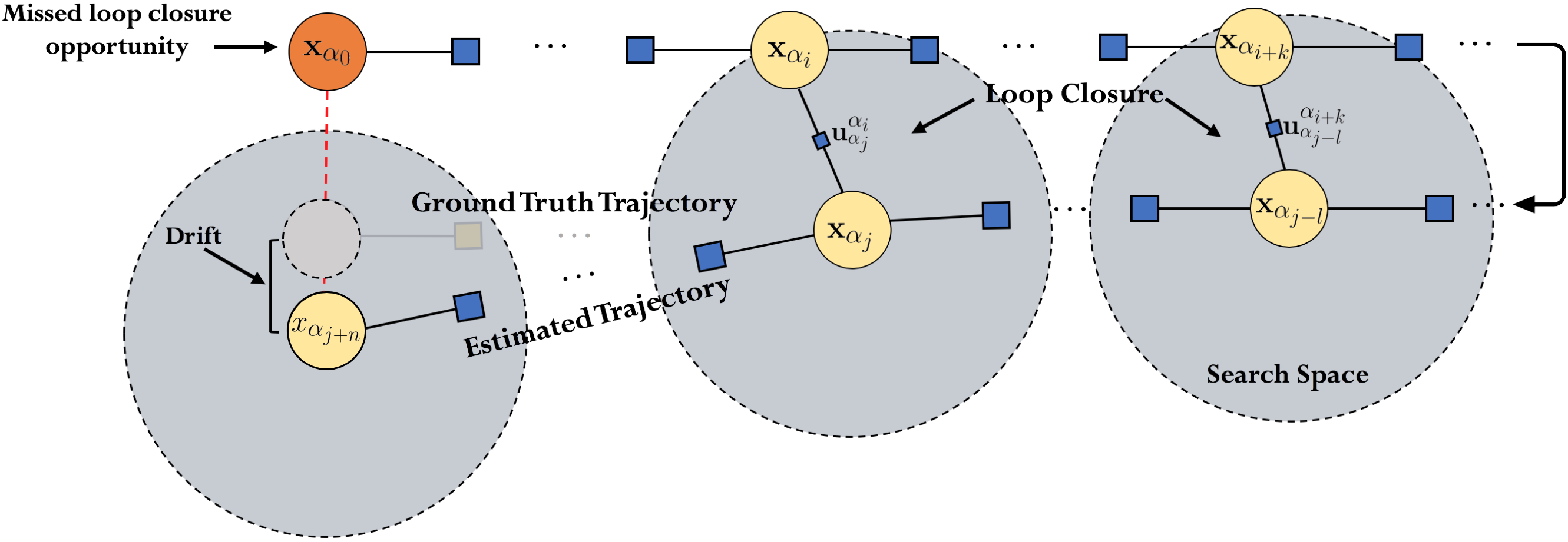}
	\caption{The search for loop closure is performed by registering a key-scan to historic key-scans that lie inside the loop closure search space. This could lead to missed loop closure opportunities if the drift in robot trajectory is larger than the expected value.}
	\label{fig:lc_search_space} 
\end{figure}
\begin{figure}[t!]
\centering
	\includegraphics[width=1.0\columnwidth, trim= 0mm 0mm 0mm 0mm, clip]{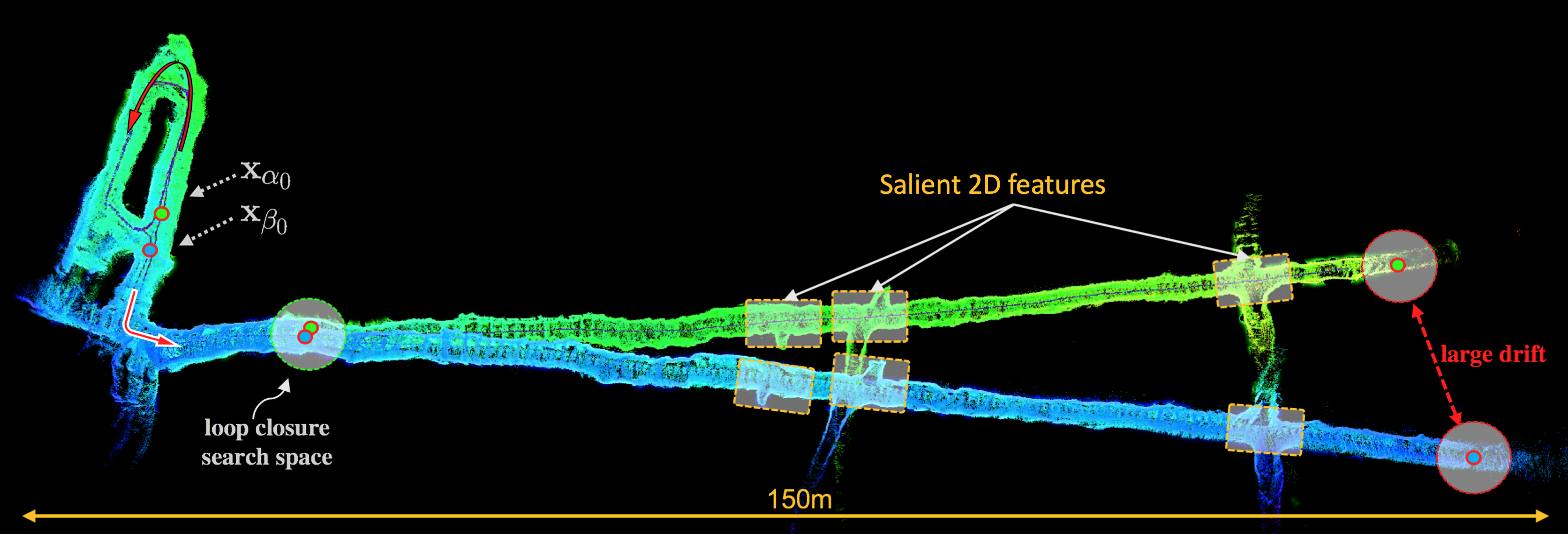}
	\caption{Partial map of the Eagle mine, Julian, CA, obtained on the base station by merging local pose graphs of two unmanned ground robots. The initial poses of the robots are shown as $\textbf{x}_{\alpha_0}$ and $\textbf{x}_{\beta_0}$. Due to significant drift in the estimated robot trajectories, loop closure opportunities are missed which has resulted in two identical tunnels as shown in blue and green, while both tunnels represent the same physical environment. The areas marked with yellow boxes show salient 2D features that can be captured in the bird's eye view of the lidar point cloud data to enable a pre-matching step for loop closure detection.}
	\label{fig:multi-robot-maps} 
\end{figure}
As discussed in Section \ref{sec:introduction}, a common method to detect loop closures in graph-based lidar SLAM is to constrain the search for loop closures to a fixed space centered at the estimated robot poses. This can be formulated as
\begin{align} \label{eq:first_lc_conditionn}
    \norm{\textbf{x}_{\alpha_i} - \textbf{x}_{\alpha_j}} < D_r,
\end{align}
where we use the positional components of robot poses $\textbf{x}_{\alpha_i}$ and $\textbf{x}_{\alpha_j}$ to compute the Euclidean distance $\norm{.}$ between two robot poses, and $D_r$ is the loop closure search radius. Although this is a reasonable constraint, in practice if the drift in robot trajectory exceeds $D_r$, loop closure opportunities will be missed.
The impact of drift is illustrated in Fig. \ref{fig:lc_search_space}, where the search  for  loop  closure  is  performed  by registering a key-scan to historic key-scans that lie inside the loop closure search space. 
Since the search space has a fixed radius, and is dependent on estimated robot poses, this could lead to missed loop closure opportunities where the drift in robot trajectory is larger than the expected value.

Fig. \ref{fig:multi-robot-maps} presents the top-down view of the distorted 3D map of the Eagle Mine located in Julian, CA, which highlights the impact of missed loop closing opportunities on the quality and consistency of constructed maps. While both branches shown in green and blue are identical as they correspond to the same physical environment, the branches are not merged due to missed loop closures.
While expanding the radius of the search space can be useful in improving loop closure detections, an expanded search space along the trajectory linearly increases the number of loop closure candidates and subsequently leads to a significant increase in the computational complexity associated with lidar scan registration. 

As shown in Fig. \ref{fig:multi-robot-maps}, avoiding missed loop closing opportunities can be achieved by expanding the search radius as needed while strategically reducing the number of nodes considered for loop closure. This not only reduces computation, but also avoids distortions of the map due to poor convergence of the ICP algorithm~\cite{local_minima} where the scene geometry does not constrain the optimization sufficiently.
In Fig. \ref{fig:multi-robot-maps}, the salient features highlighted with yellow boxes can be used to develop a drift-resilient pre-matching step as described in the next section.

\subsubsection{Saliency-Based Geometric Loop Closing} \label{Sec:semantic_lc}
\begin{figure}[b!]
\centering
	\includegraphics[width=1.0\columnwidth, trim= 0mm 0mm 0mm 0mm, clip]{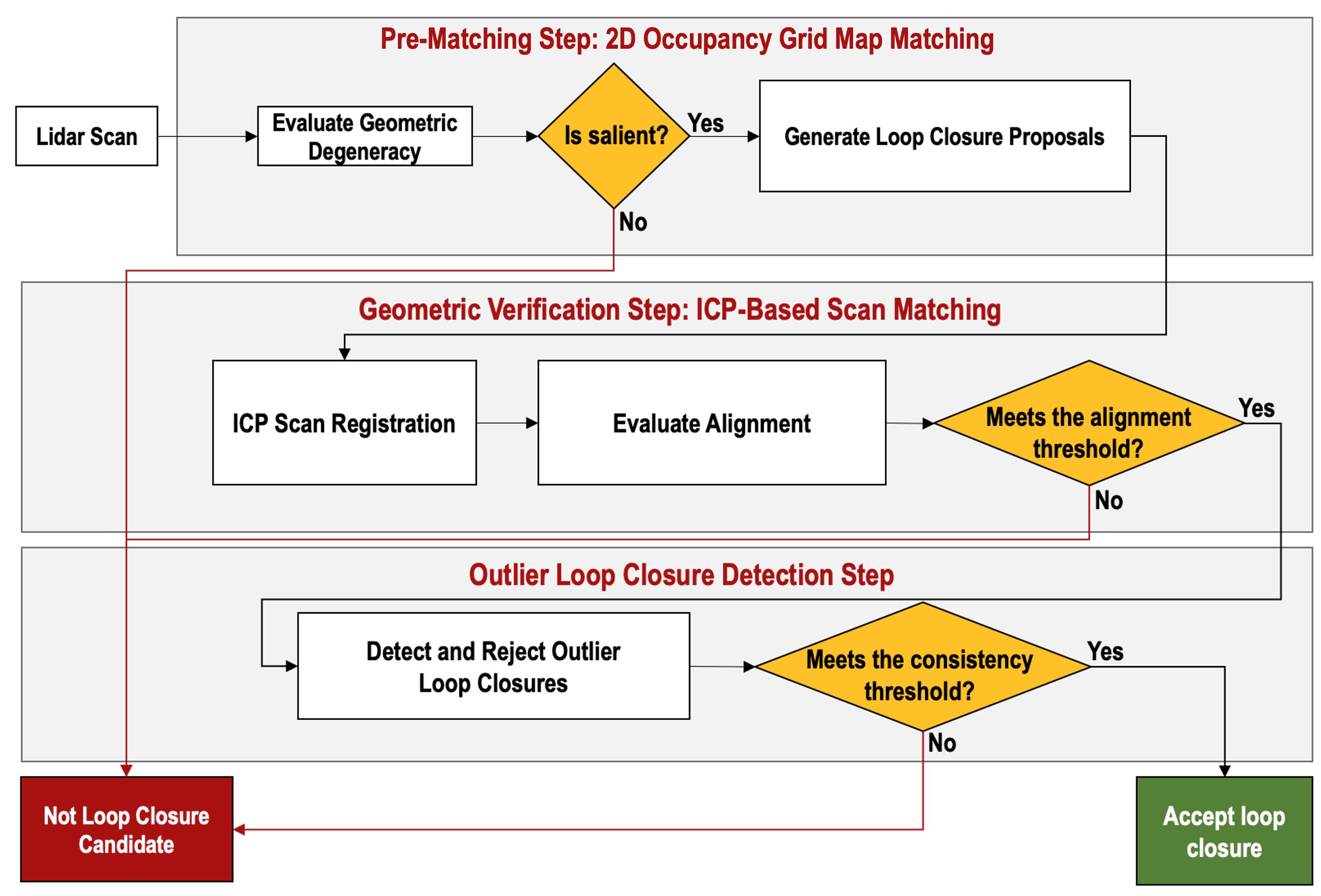}
	\caption{An overview of the saliency-based loop closure detection method.}
	\label{fig:loop_closure_diagram} 
\end{figure}
In this section, we describe the proposed multi-stage loop closing method for a single-robot system, and then show how it can be extended to a multi-robot SLAM system to enable map alignment and merging on a based station in a centralized SLAM architecture.
Fig. \ref{fig:loop_closure_diagram} provides an overview of the proposed multi-stage pipeline.
The method consists of three main layers, namely pre-matching, geometric verification and outlier rejection.
In the pre-matching step, 2D occupancy grid maps constructed from key-scans are used for assessment and evaluation of potential loop closures over segments of the mapped environment that are determined to be fully observable.
In contrast to the BGLC method that relies on a fixed search radius in the local neighborhood of each robot pose, the pre-matching step has the key advantage of enabling a fast and global screening of putative loop closures, and thus, it is pose-invariant. 
The pre-matching step is based on 2D occupancy grid map matching. The essence of using 2D occupancy grid maps is to capture the spatial configuration of the local scenes by creating bird's eye views of the lidar point cloud data. This enables the search for loop closures by performing occupancy grid map matching, where the objective is to find scenes with similar spatial configuration.

2D occupancy grid maps were first introduced by A. Elfes in 1985~\cite{Elfes}, and have evolved over time~\cite{ConfidenceRichMapping} as one of the most common types of map used in robot navigation and path planning. 
In this work, each occupancy grid map has $250 \times 250$ cells, and corresponds to an area of $5$ m $\times$ $5$ m in the physical environment. Each cell stores a corresponding occupancy belief $b_i(x,y)_{\;0 \leq x,y < 250}$, representing the estimated probability that a static or stationary object is present in that cell location.

In order to construct an occupancy grid map from a key-scan, the ROS 2D costmap tool \cite{Costmap} is used to obtain a slice of robot's surrounding 3D world from lidar point cloud data. This is achieved by filtering the point cloud to remove the points that comprise the ground plane, as well as the points that appear higher than the highest point on the robot. The filtered point cloud is then projected onto a $XY$ plane to construct the occupancy grid map. 
A cell is considered as being in one of three possible states: free for $b_i(x,y) < 0.5$, unknown for $b_i(x,y) = 0.5$, and occupied for $b_i(x,y) > 0.5$.
We convert the belief map $b_i(x,y)$ to a binary map $o_i(x,y)$ using
\begin{equation}
o_i(x,y)=
\begin{cases}
  0 & \text{if} \ b_i(x,y) < 0.5 \\
  1 & \text{if} \ b_i(x,y) \geq 0.5
\end{cases},
\end{equation}
where $o_i(x,y)$ can be interpreted as a 2D binary map image of the local scene.

\subsubsection{Visual saliency in occupancy grid maps}
\begin{figure}[b!]
\centering 
  \includegraphics[width=1\columnwidth]{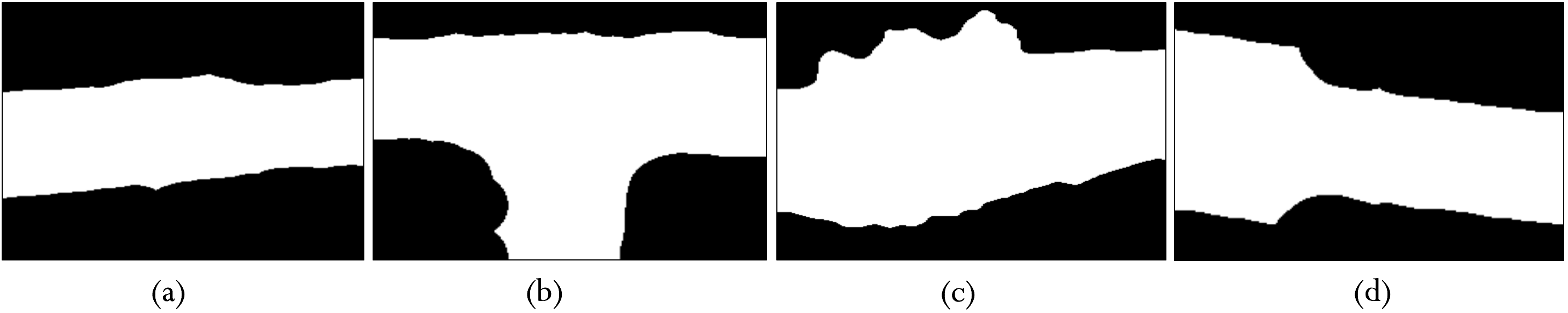}
  \caption{(a) shows an occupancy grid map with lack of visual saliency. (b), (c) and (d) show occupancy grid maps with salient visual features.} \label{fig:costmaps}
\end{figure}

While loop closure detection is a crucial requirement in single- and multi-robot SLAM systems, it is equally crucial to avoid closing loops in ambiguous areas with high level of geometric degeneracy.
Attempting loop closures in these areas can lead to detection of false or weak loop closures that can result in catastrophic distortions of the map.
In a geometrically degenerate scene like a long and featureless tunnel with flat walls, the spatial configuration of the scene captured in 2D occupancy grid maps also lack distinctive features sufficient for localization, a condition that we refer to it as \emph{lack of visual saliency}. 
Fig. \ref{fig:costmaps} presents examples of occupancy grid maps constructed from lidar scans in an underground tunnel. 
While Fig. \ref{fig:costmaps}-(a) can lead to perceptual aliasing and data association ambiguity due to lack of visual saliency, Fig. \ref{fig:costmaps}-(b-c-d) show maps with sufficiently distinctive spatial configurations that can be used for global localization in the pre-matching step.
By relying on the degeneracy metric $\kappa$ in (\ref{Eq:condition_number}), the level of observability of the scene can be analyzed in real-time to remove ambiguous regions as shown in Fig. \ref{fig:costmaps}-(a) from loop closure consideration. In the next section, we develop metrics to determine putative loop closures based on occupancy grid map matching.

\subsubsection{Pre-matching step: 2D occupancy grid map matching}
The search for loop closures can be formulated as a place recognition problem where a robot establishes global localization by registering a salient occupancy grid map obtained at its current pose to a set of salient occupancy grid maps obtained in the past throughout its trajectory. This can be interpreted as a special instance of 2D image registration problem where occupancy grid maps are viewed as 2D binary map images. 

Let $o_{\alpha_i}$ and $o_{\alpha_j}$ denote two map images obtained from key-scans $z_{\alpha_i}$ and $z_{\alpha_j}$ of robot $\alpha$. 
From (\ref{Eq:two_stage}) $LC^s(\alpha_i, \alpha_j)$ is the pre-matching step, which evaluates the level of similarity between a pair of occupancy grid maps based on feature matching.
In order to register two map images, the set of all salient 2D features are extracted from $o_{\alpha_i}$ and $o_{\alpha_j}$. 
In this paper, the Oriented FAST and Rotated BRIEF (ORB) features by Ethan Rublee et al.~\cite{ORB}, is used over other methods (i.e., SIFT, SURF) due to its key qualities: (i) outstanding speed and performance, (ii) resistance to image noise, (iii) and rotation invariance.
We find the set of $N_{corr}$ corresponding features by using the Fast Library for Approximate Nearest Neighbors (FLANN) presented by Chanop et al.~\cite{Silpa}. FLANN is a fast local approximate nearest neighbors method that is commonly used to match keypoints found between corresponding images and to compute the set of corresponding feature points.
Finally, using the Random Sample Consensus (RANSAC) algorithm~\cite{RANSAC}, a homography $\hat{\textbf{M}}^{o_{\alpha_i}}_{o_{\alpha_j}}$ that best describes the 2D geometric transformation between the occupancy grid maps is computed. 

The objective of the pre-matching step is to identify the most likely loop closure candidates along robot's trajectory based on spatial similarity of the scenes.
The correspondence confidence,
\begin{align} \label{Eq:CC}
    \zeta_{i,j} = \frac{N_{in}}{N_{corr}},
\end{align}
which is the ratio of the inliers $N_{in}$, to the total number of corresponding points $N_{corr}$, is often used as a fitness measure in feature-based image registration. In our experiments, if the number of inliers is less than a threshold (i.e., $N_{in} \leq 20$), the map image is removed from loop closure consideration. In a best case scenario, where all corresponding points are inliers, the correspondence confidence $\zeta_{i,j} = 1$.

In perceptually-degraded environments, we can encounter situations where after finding the set of inliers using RANSAC, the correspondence confidence score is high even though the match is incorrect. In general, the theoretical breakdown point of all robust estimators, where there is no general guarantee of success in detection of true inliers, is when the percentage of outliers is more than $50\%$ \cite{RANSAC_outlier_rejection}. As a result, based on the level of self-similarity and ambiguity between map images, the number of selected inliers and accuracy of the estimated homography can vary.
In order to reduce perceptual aliasing and data association ambiguity, we introduce the transformation confidence to identify high values of correspondence confidence scores that are unreliable. 
Let $f_k^{o_{\alpha_i}}$ and $f_k^{o_{\alpha_j}}$ denote the $k$th pair of corresponding 2D features in map images $o_{\alpha_i}$ and $o_{\alpha_j}$, where each feature is defined with its 2D image coordinate.
Given the computed homography from RANSAC, the residual error between the $k$th pair of corresponding feature points is given by
\begin{align} \label{eq:residual_feature_error}
    r_k =  \norm{\hat{\textbf{M}}f_k^{o_{\alpha_i}} - f_k^{o_{\alpha_j}}}
\end{align}
where $r_k$ is the Euclidean distance between the $k$th pair of aligned corresponding points. 
Using (\ref{eq:residual_feature_error}), the mean squared error for the set of all $N_k$ corresponding points is given by
\begin{align} \label{MSE_image_Features}
    \epsilon_{i,j} =\frac{1}{N_k}\sum_{k=1}^{N_k}\norm{r_k}^2.
\end{align}

Using the computed mean squared error $\epsilon_{i,j}$, we evaluate the quality of the computed homography between $o_{\alpha_i}$ and $o_{\alpha_j}$ by computing
\begin{align} \label{Eq:TC}
    \Lambda_{i,j} = \frac{1}{1 + \epsilon_{i,j}},
\end{align}
where $\Lambda_{i,j}$ is the \emph{transformation confidence} score which is close to $1.0$ for a perfect match, and close to zero for a false match with a large number of outlier correspondences.
This is used to reject cases which have a high correspondence confidence but also a high matching error by defining the \emph{similarity confidence} metric denoted by $\Psi_{i,j}$, that is obtained from the product of the correspondence and transformation confidence scores as given by
\begin{align} \label{Eq: similarity_score} 
    \Psi_{i,j} = \zeta_{i,j} \cdot \Lambda_{i,j}.
\end{align}

\begin{figure}[b!]   
\centering 
	\includegraphics[width=1.0\columnwidth]{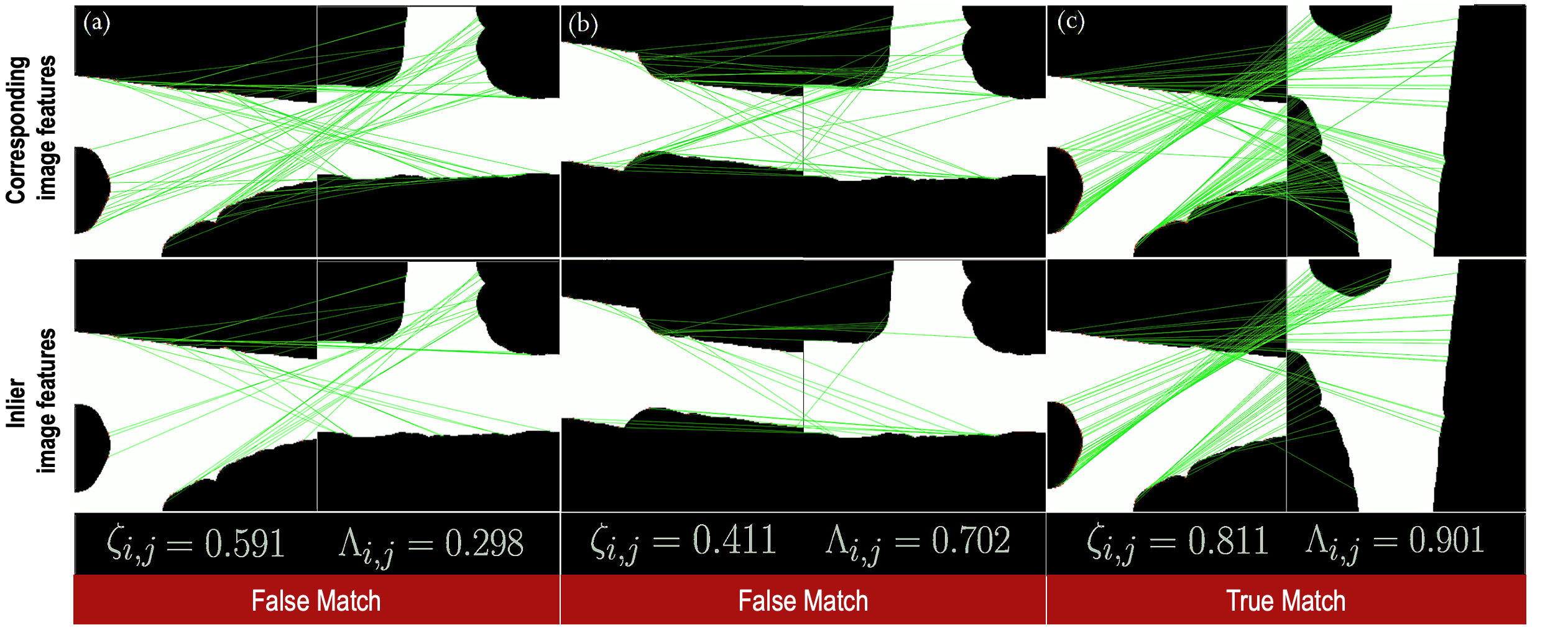}
	\caption{The top and bottom rows present the set of corresponding and inlier features for an occupancy grid map that is matched against 3 salient occupancy grid maps. The correspondence and transformation confidence scores are computed for each pair. Only (c) shows a true positive match.}
	\label{fig:grid_matching}
\end{figure}
\begin{figure}[b!]
\centering 
	\includegraphics[width=1.0\columnwidth]{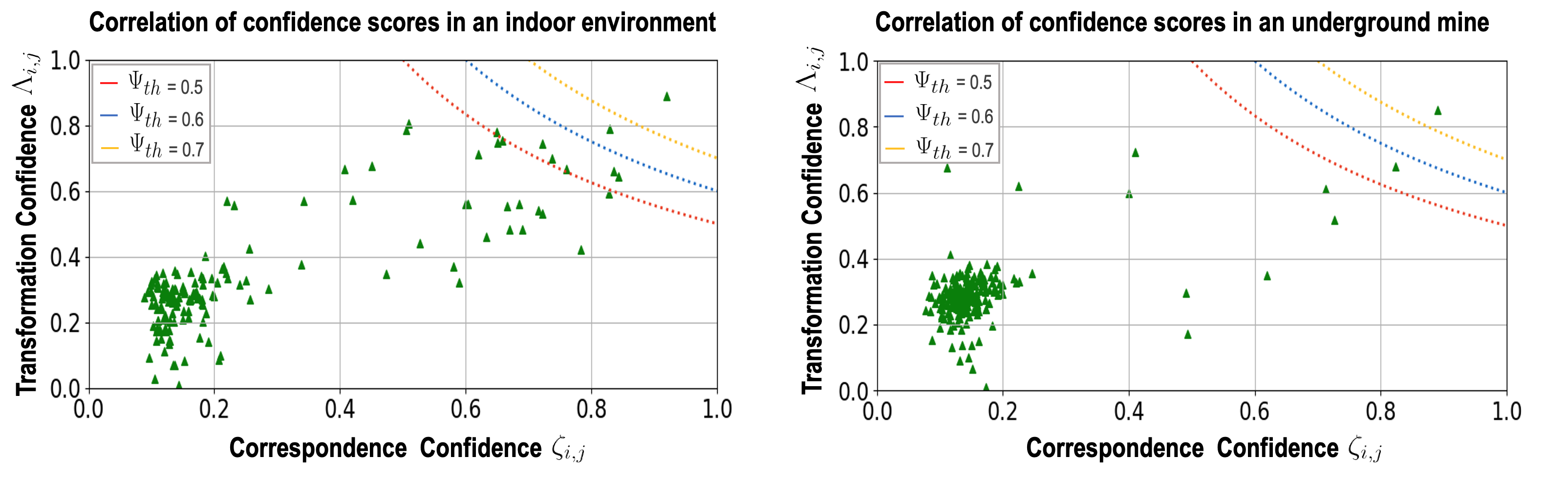}
	\caption{2D scatter plots showing the correspondence and transformation confidence scores for occupancy grid map matching in an indoor office environment and an underground mine.}
	\label{fig:match_confidence}
\end{figure}
If the similarity confidence score is larger than a threshold, occupancy grid maps $o_{\alpha_i}$ and $o_{\alpha_j}$ are considered as a loop closure candidate. For true matches where the correspondence confidence scores are greater than $0.7$, the transformation confidence scores also tend to be large.
Fig. \ref{fig:grid_matching} presents some representative examples of occupancy grid map matching where the set of corresponding and inlier features are visualized.
Fig. \ref{fig:grid_matching}-(a-b) present the correspondence and transformation confidence scores for two false image matches. In Fig. \ref{fig:grid_matching}-(a) despite of having a large fraction of inlier features, the computed transformation confidence is small indicating the poor quality of the match due to false set of inlier features. 
Fig. \ref{fig:grid_matching}-(b) presents an opposite scenario where the number of inlier features is relatively small, while the computed transformation confidence score is large. This is mainly because most of the features are concentrated in patches at certain areas in the image resulting in a low alignment error. 
Fig. \ref{fig:grid_matching}-(c) presents a true match where both correspondence and transformation confidence scores are large. The results illustrate that while the individual correspondence and transformation confidence scores do not have enough discrimination power to determine a true loop closure, the product of both metrics can be used as a reliable similarity confidence metric to identify most similar matches.

Fig. \ref{fig:match_confidence} presents 2D scatter plots for correspondence and transformation confidence scores in an indoor office environment and an underground mine where $200$ occupancy grid maps in each environment are evaluated for loop closure. 
On both plots, contours of similarity threshold values $\Psi_{th}$ are shown. In the underground mine, a correspondence confidence threshold of $\zeta_{i,j} = 0.8$ would select the same loop closure candidates as similarity score $\Psi_{th} = 0.7$. Moreover, there are very few matches with correspondence scores $\zeta_{i,j} \geq 0.5$ that are all rejected by using the similarity threshold $\Psi_{th}$. However, for the indoor office environment with self-similar office cubicles, a much larger number of candidate loop closures with correspondence scores $\zeta_{i,j} \geq 0.5$ are rejected by using the similarity confidence threshold $\Psi_{th}$.


\subsubsection{Geometric verification step}
Let ${\alpha_i}$, ${\alpha_j}$ and $\hat{\textbf{M}}^{o_{\alpha_i}}_{o_{\alpha_j}}$ be the candidate loop closure key-nodes in the pose graph, and the computed homography obtained from the pre-matching step. As presented in Fig. \ref{fig:loop_closure_diagram}, we use a geometric verification step of $LC^g(LC^s(\alpha_i, \alpha_j))$ in (\ref{Eq:two_stage}), to verify the quality of the loop closure candidate using the ICP-based scan matching shown in (\ref{Eq:scan-to-scan}) to align the corresponding key-scans.
The performance of the ICP algorithm relies heavily on the quality of initialization; with a poor initial guess the algorithm is susceptible to local minima, especially if the actual 3D motion between two lidar scans is large. In order to improve convergence of the ICP algorithm to the optimal solution in the geometric verification step, the algorithm is seeded using the obtained 2D rotation matrix from the computed homography $\hat{\textbf{M}}^{o_{\alpha_i}}_{o_{\alpha_j}}$, where a 3D transformation matrix with a $Z$-axis rotation and zero translation is constructed to initialize the ICP algorithm. This can be interpreted as a two-stage optimization process where the objective is to refine the computed geometric transformation obtained from pre-matching step to find the best estimate of the relative 3D motion between two robot poses.
After aligning the point clouds, the convergence criteria is evaluated using the alignment error $\mathcal{E}(.)$ from (\ref{Eq:MSE}) before sending the loop closure constraint to the SLAM back-end for outlier rejection verification. 

\subsubsection{Outlier rejection step} \label{sec:outlier_rejection}
The SLAM back-end is implemented using the Georgia Tech Smoothing and Mapping (GTSAM)~\cite{GTSAM} library that implements smoothing and mapping using factor graphs and Bayes networks as the underlying computing paradigm.
\begin{figure}[b!]    
\centering 
	\includegraphics[width=1.0\columnwidth]{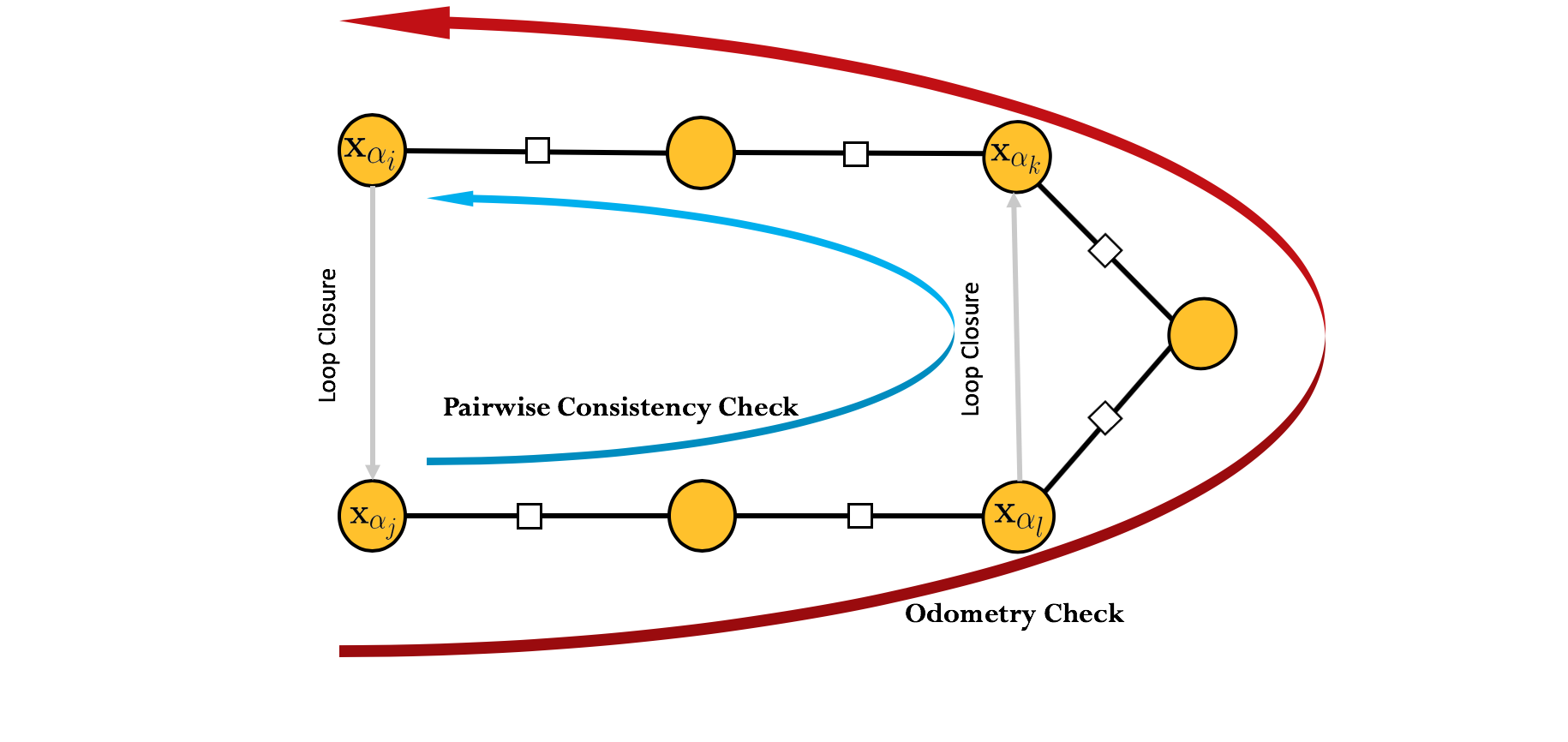}
	\caption{Outlier rejection is performed by evaluating the quality of a loop closures in terms of consistency with odometry edges as well as pairwise consistency with previously closed loops in the pose graph.}
	\label{fig:pcm}
\end{figure}
Since pose graph optimization relying on least square optimization methods is not robust against outliers, the back-end relies on an outlier loop closure detection method as illustrated in Fig. \ref{fig:pcm} to prevent the optimization from producing incorrect solutions when the front-end produces spurious loop closures.
The outlier loop closure detection is based on the Pairwise Consistent Measurement Set Maximization (PCM) method presented in \cite{LAMP, Door_SLAM}.
Once the SLAM back-end receives putative loop closures from the geometric verification step, the quality of loop closures is evaluated both in terms of consistency with odometry edges, as well as pairwise consistency with previous loop closures. 

As illustrated in Fig. \ref{fig:pcm}, assuming a negligible measurement noise, when a loop is closed and an edge is added between two key-nodes in the pose graph, accumulation of the loop closure edge with all odometric or loop closure edges along the cycle will compose to the identity~\cite{Carlone14}. Using this basic observation, back-end rejects loop closures detections that lead to a large transformation error along the cycles in the pose graph.
Once a loop closure is confirmed and a new constraint is added between two key-nodes, the pose graph is optimized using iterative nonlinear optimization methods, (e.g. levenberg-marquardt) to obtain the best estimate of robot poses from the set of all constraints. Algorithm $1$ summarizes the multi-stage loop closure detection process.

\begin{algorithm}[t!]
\caption{Saliency-Based Geometric Loop Closing (SGLC)}\label{euclid}
\hspace*{\algorithmicindent} \textbf{Input : a query lidar scan $z_{i}$} \\
\hspace*{\algorithmicindent} \textbf{Input : a set of previously obtained lidar scans $Z$} \\
\hspace*{\algorithmicindent} \textbf{Output: a loop closure detection}
\begin{algorithmic}[1]
\Procedure{FindLoopClosure}{$Z$, $z_i$}
\If{($log(\kappa_i) \leq \kappa_{th}$)} \Comment{Eq. \ref{Eq:condition_number}}
    \State $o_i \gets \Call{ConstructOccupancyGridMap}{z_i}$
    \For {($\forall  z_{j} \in Z$)}
        \If{($log(\kappa_j) \leq \kappa_{th}$)}
        \State $o_j \gets \Call{ConstructOccupancyGridMap}{z_j}$
            \If {$(\Call{Prematch}{o_{i}, o_{j})}$}  \Comment{Pre-matching step}
                \State \Return $\textbf{M}^i_j$
                \If {(\Call{GeometricVerification}{${z_i, z_j, \textbf{M}^i_j}$})}     \Comment{Geometric Verification step}
                \State \Return $\textbf{u}^i_j$
                    \If {(\Call{NotAnOutlier}{$\textbf{u}^i_j$})}   \Comment{Outlier rejection step}
                       \State \Return $\textbf{U} \gets \textbf{U} \cup \textbf{u}^i_j$ \Comment{Add new loop closure to the set of constraints}
                    \EndIf
                \EndIf
            \EndIf
        \EndIf
    \EndFor
\EndIf
\EndProcedure
\end{algorithmic}
\end{algorithm}
\begin{table}[b!]
\centering
\caption{List of the explored underground mines in our field experiments.}
\label{tab:mines}
\begin{tabular}{|M{4.0cm}|M{2.2cm}|M{1.8cm}|M{2.0cm}|}\hline
Name of the mine & Autonomously Traversed Distance     & Type of mine    & Location \\ \hline
Arch Pocahontas Mine   & 1100 m & Coal Mine &  Beckley, WV \\ \hline
Beckley Exhibition Mine & 1000 m & Coal Mine &  Beckley, WV \\ \hline
Bruceton Safety Research Mine  & 1400 m & Coal Mine & Pittsburgh,  PA \\ \hline
Bruceton Experimental Mine  & 700 m & Coal Mine & Pittsburgh,  PA \\ \hline
Highland Mine & 1400 m & Coal Mine &  Logan, WV  \\ \hline
Eagle Mine    & 500 m & Gold Mine & Julian, CA \\ \hline
\end{tabular}
\end{table}

\subsubsection{Multi-Robot collaborative mapping}
Assuming the initial positions of robots are known, and there is at least a partial overlap between robot trajectories, map merging is performed on the base station by finding the correspondences between the pose graphs. This is achieved using the same multi-stage loop closure detection process used in the single-robot scenario, where during the pre-matching step the search is constrained to the most observable areas in the local pose graphs to identify putative loop closures. Upon determination of inter-robot loop closure candidates, each candidate undergoes geometric verification and outlier rejection steps before performing a global pose graph optimization to find the most probable configuration of the nodes in the merged pose graphs given the set of all intra- and inter-robot constraints.

\section{Experiments} \label{sec:Experiments}
In this section, we present performance and computational analysis of our proposed saliency-based geometric loop closing (SGLC) method in a variety of complex underground environments listed in Table \ref{tab:mines}. Moreover, we provide quantitative and qualitative comparison of the localization and mapping results against the basic geometric loop closing (BGLC) method, as well as the LeGO-LOAM that is a state-of-the-art SLAM method with loop closure detection capability that is recently introduced by Shan et al.~\cite{LeGoLOAM}.

\begin{figure}[b!]
\centering 
  \includegraphics[width=1\columnwidth]{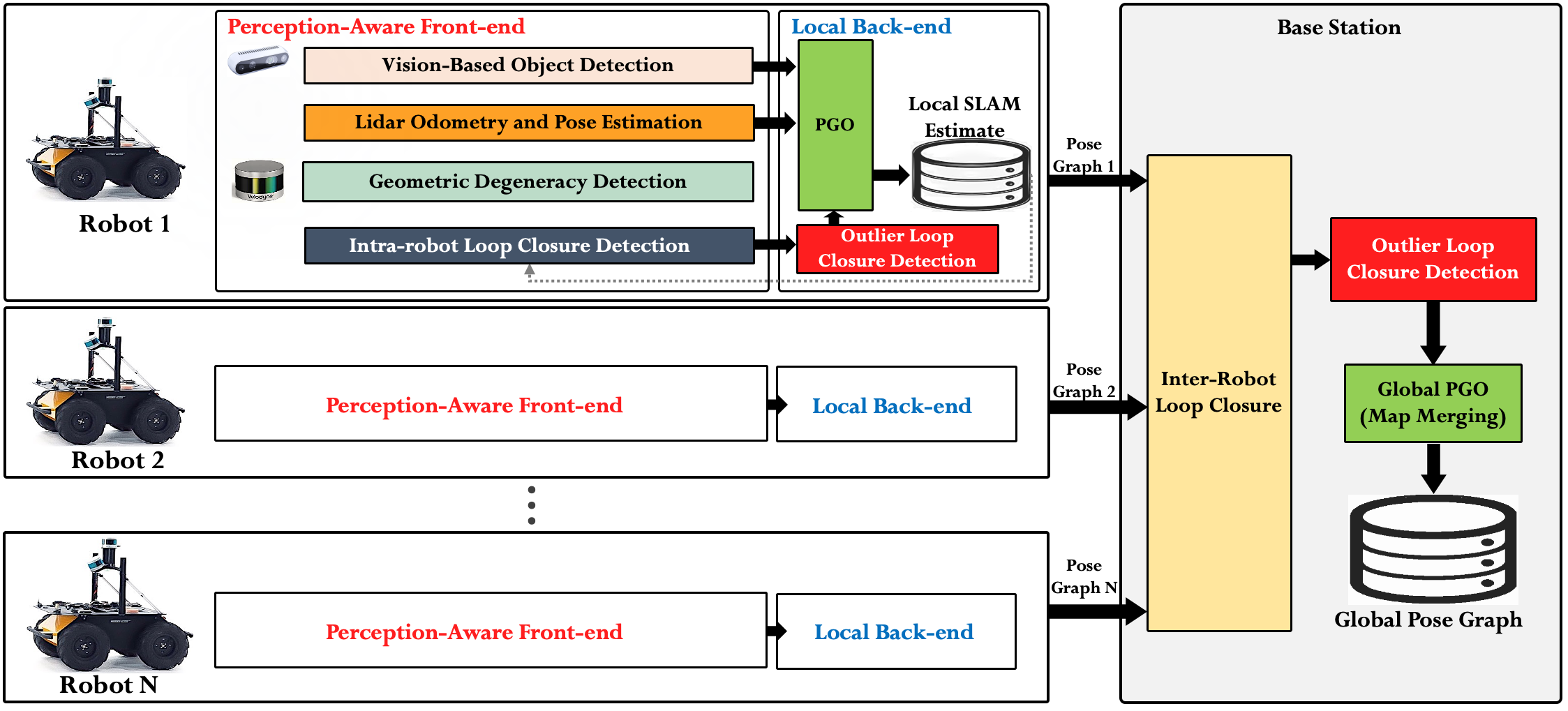}
  \caption{Centralized multi-robot collaborative mapping architecture. \label{fig:multiArchitecture}}
\end{figure}
\begin{figure}[b!]    
\centering 
	\includegraphics[width=1.0\columnwidth]{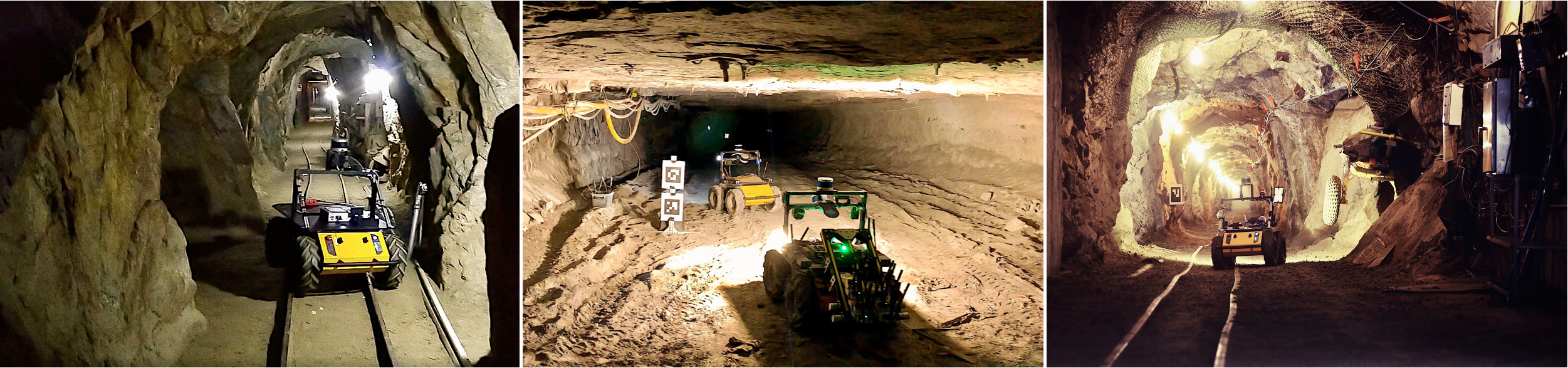}
	\caption{Examples of the sensor-degraded underground environments explored by autonomous ground robots (Husky A200). From left to right, Eagle Mine in Julian, CA, Arch Pocahontas Mine in Beckley, WV and Edgar Mine in Idaho Springs, CO.}
	\label{fig:test_environments}
\end{figure}

As presented in Fig.~\ref{fig:multiArchitecture}, we use a centralized multi-robot SLAM architecture, developed in the context of DARPA Subterranean Challenge, where the main objective of this robotic challenge is to explore and map unknown and extreme underground environments as shown in Fig. \ref{fig:test_environments}, using autonomous single- or multi-robot systems. 
When a robot exploring the unknown environment is within the wireless communication range, it communicates its pose graph and the constructed map to a base station. 
The base station is an Intel Hades Canyon NUC8i7HVKVA ($4\times1.9$ GHz, $32$ GB RAM), that is responsible for receiving and merging the local pose graphs of individual robots to build a consistent global map of the unknown environment. 
In the context of this Challenge, the base station does not communicate the constructed global maps back to the robots and thus, a real-time performance is not strictly required. However, the constructed global maps can be used for reliable navigation of robots in the future. Hence, the proposed architecture can be scaled to larger robot teams.

Each robot is a Husky-A200 series that relies on a VLP-16 Puck Lite lidar scanner~\cite{PUCK} and an Intel NUC 7i7DNBE  ($4\times1.9$ GHz, 32 GB RAM) processor for onboard simultaneous localization and mapping. 
By relying on an onboard Intel RealSense D435 RGB-D camera, each robot uses vision-based object detection based on the You Only Look Once (YOLO)~\cite{YOLO} algorithm to detect, classify and localize objects of interest in the environment.

As obtaining the ground truth data in large-scale underground environments is a challenging task, similar to our method in LAMP \cite{LAMP}, we obtain a proxy for the ground truth trajectories by enforcing the ground truth locations of the known objects and fiducial markers (as shown in Fig. \ref{fig:test_environments}) in the best pose graph of each robot, and use the resulting optimized trajectory as ground truth.

\subsection{Performance analysis of pre-matching step} 
We evaluate the sensitivity and specificity of the similarity confidence metric used in the pre-matching step in (\ref{Eq: similarity_score}) in five underground environments by using ROC analysis.
\begin{figure}[b!]   
\centering 
	\includegraphics[width=1\columnwidth]{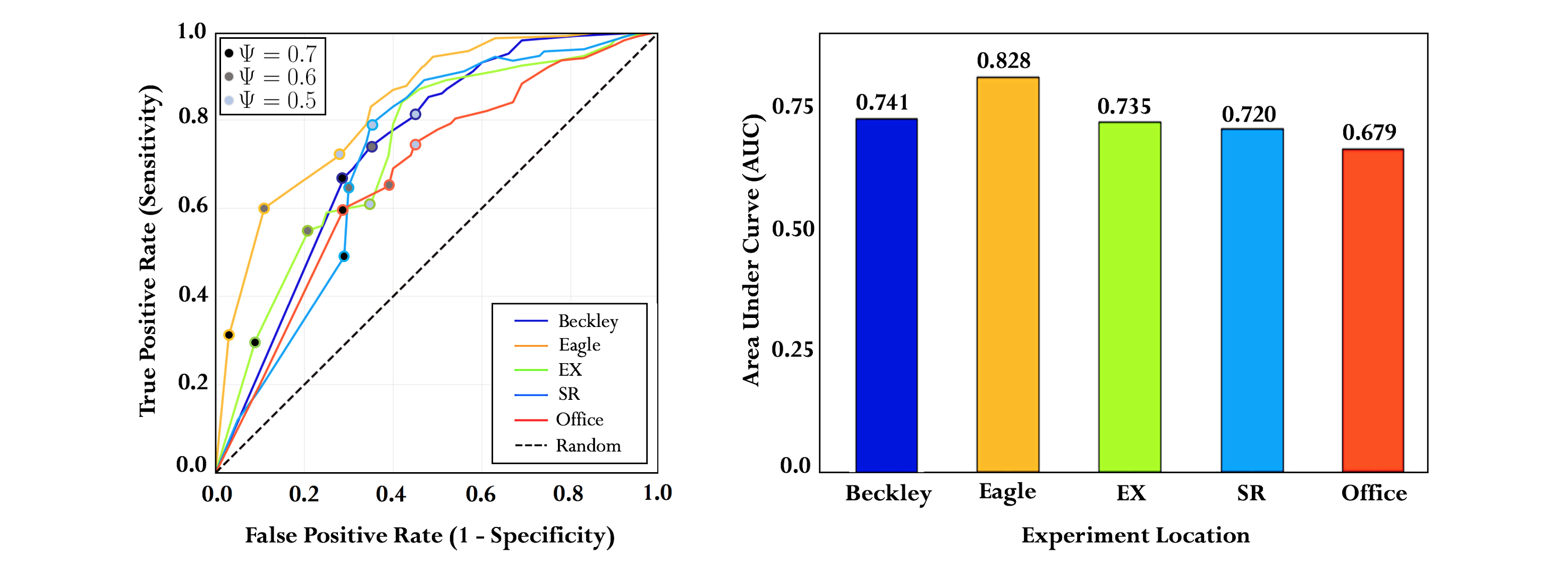}
	\caption{The ROC curves and AUC for pre-matching step in five underground environments created by plotting the true positive rate (TPR) against the false positive rate (FPR) at various similarity confidence $\Psi$  threshold settings.} \label{fig:semantic_lc_analysis}
\end{figure}
In each environment, a set of $100$ salient occupancy grid maps are obtained from the key-scans, where $20$ of the occupancy grid maps in each environment correspond to previously visited locations and represent true loop closures. Fig. \ref{fig:semantic_lc_analysis} shows that loop closures can be reliably detected in various environments by relying on occupancy grid map matching where an average $\text{AUC} = 0.756$ is achieved in correctly identifying loop closures among all environments. Different threshold values are marked on each plot to show low sensitivity of loop closures to threshold values across five different underground environments.

While the indoor office represents a structured environment with salient geometric features, accurate place recognition used in the pre-matching step in this environment is shown to be more challenging as compared to the underground mines. This is mainly due to perceptual aliasing and data association ambiguity that arises from the \emph{self-similarity} of spatial configuration of local scenes in the indoor office environment. The presence of multiple office cubicles with identical geometric structures leads to larger number of false positive detections in the pre-matching step. Underground tunnels have fewer repetitive geometric structures which contributes to the reduction of perceptual aliasing in the pre-matching step. This highlights the importance of using a geometric verification step, and a back-end equipped with an outlier rejection capability to verify each loop closure candidate in terms of accuracy and consistency before adding it to the pose graph.

\begin{figure}[b!]   
\centering 
	\includegraphics[width=1\columnwidth]{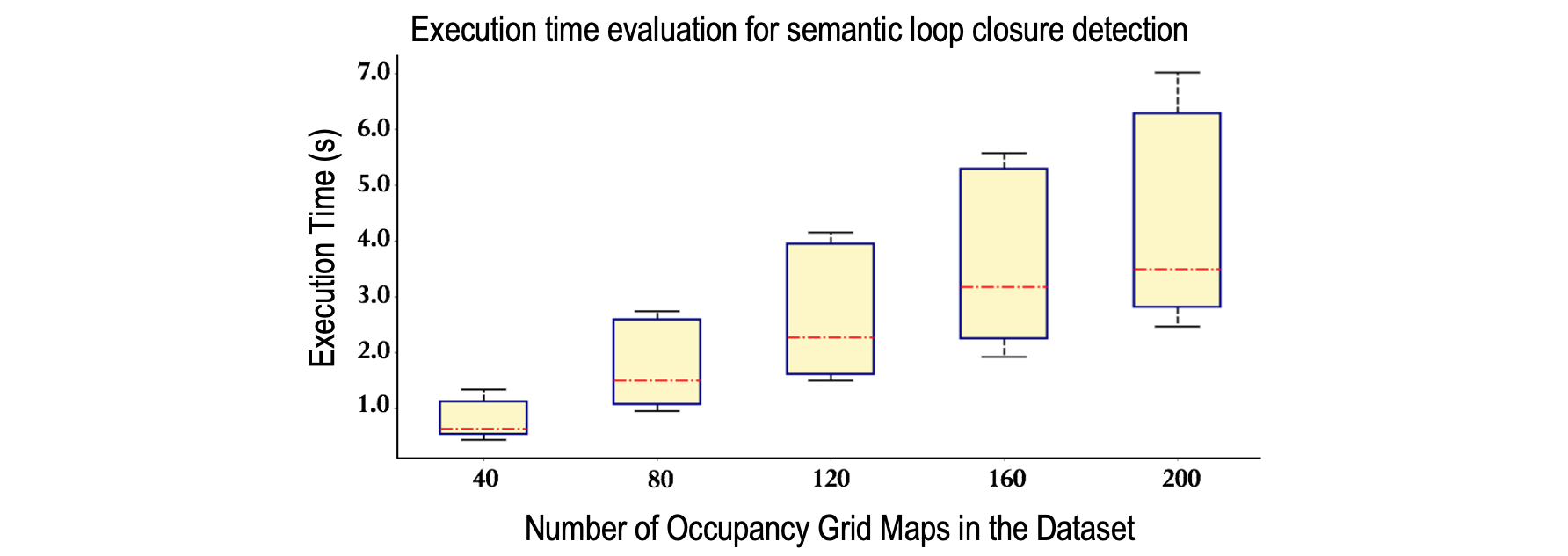}
	\caption{Execution time analysis for the pre-matching step. Each box comprises the computation time values ranging from the first to the third quartile. The median is indicated by the dashed red horizontal bar. The whiskers extend to the farthest data points that are within $1.5$ times the interquartile range.} 	\label{fig:computation_time}
\end{figure}
\begin{figure}[b!]   
	\includegraphics[width=1.0\columnwidth]{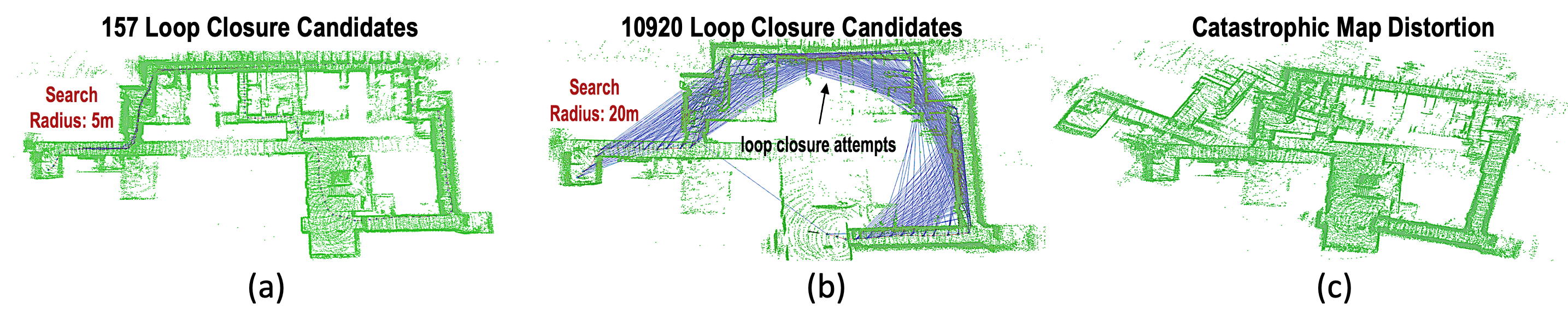}
	\caption{(a) Expanding the loop closure search radius from $5$m in (a) to $20$m in (b) based on the BGLC method dramatically increases number of attempted loop closures. The blue lines show all node pairs in the pose graph that are considered for loop closure. This in turn, increases the probability of spurious loop closures that can result in catastrophic distortions of the map as shown in (c).}
	\label{fig:BGLC-expanded-radius}
\end{figure}

\subsection{Computation analysis of pre-matching step} 
In this experiment, we evaluate the execution time of the pre-matching and geometric verification steps in order to characterize the computational complexity introduced by the pre-matching step. In each experiment, a set of $10$ occupancy grid maps are used as query map images. Using the pre-matching process described in Section \ref{Sec:semantic_lc}, each query map image is registered to all map images in the corresponding dataset to obtain the similarity confidence scores that describe the level of visual similarity between the query image and all map images in the dataset. 
Fig. \ref{fig:computation_time} shows the execution time for loop closure detection for datasets of different sizes. The box plots show the average execution time does not grow linearly as the size of the datasets increase. This is due to the fact that map images with a small number of corresponding features (less than $20$ in our experiments) are discarded from loop closure consideration and their similarity confidence score is set to zero before further processing.

As presented in Section \ref{sec:lidarLC}, considering the significant number of points in each lidar scan, if the search space in basic geometric loop closing method is expanded to include all nodes in the graph, this results in the quadratic computational complexity $O(n^2)$ \cite{ICPcomplexity} of the ICP algorithm, incurring the BGLC method a prohibitive computational complexity cost of $O(n^3)$ in detection of loop closures for $n$ nodes in the pose graph.
Fig. \ref{fig:BGLC-expanded-radius}-(a-b) show maps obtained in an indoor office environment by using the BGLC method, where two search radii of $5$m and $20$m are used for detection of loop closures. As illustrated with blue lines between loop closure candidates, increasing the search radius results in a dramatic increase ($10920$ loop closure candidates) in the number of attempted loop closures. This not only increases the computational load associated with ICP-based lidar scan registration, but also increases the probability of spurious loop closures as shown in the map in Fig. \ref{fig:BGLC-expanded-radius}-(c).

In our proposed SGLC method, the pre-matching step relies on a pair-wise occupancy grid map registration process with $O(n)$ complexity \cite{FMcomplexity}. 
Moreover, the determination of geometric degeneracy helps to reduce the number qualifying nodes by constraining the search to areas with full observability. In a feature-rich environment where all nodes qualify for the pre-matching step, the algorithm will incur a computational complexity cost of $O(n^2)$ for the set of all $n$ binary map images which has a lower complexity than the BGLC method and can be executed on a separate thread to avoid affecting the real-time performance of the front-end.

\subsection{Large-scale mapping using SGLC method} 
In this section, we provide analysis of localization and mapping accuracy and comparison with the state-of-the-art in a verity of large-scale and extreme unknown environments. In our earlier work LAMP ~\cite{LAMP}, a capability was developed to allow the operator to manually add loop closure constraints to the pose graph if loop closures were missed due to large drift in the estimated robot trajectory. While this is an effective tool to reduce the drift and improve the quality of the constructed maps, it requires a human in the loop and thus, is unreliable in large-scale and complex underground environments with intermittent or regional wireless connections between the operator and the robots.
In the experiments presented in this section, our proposed SGLC method removes the human from the loop and enables an automatic detection of loop closures that would otherwise be missed due to accumulation of errors in robot trajectories.
\begin{figure}[b!]   
  \includegraphics[width=1.0\columnwidth]{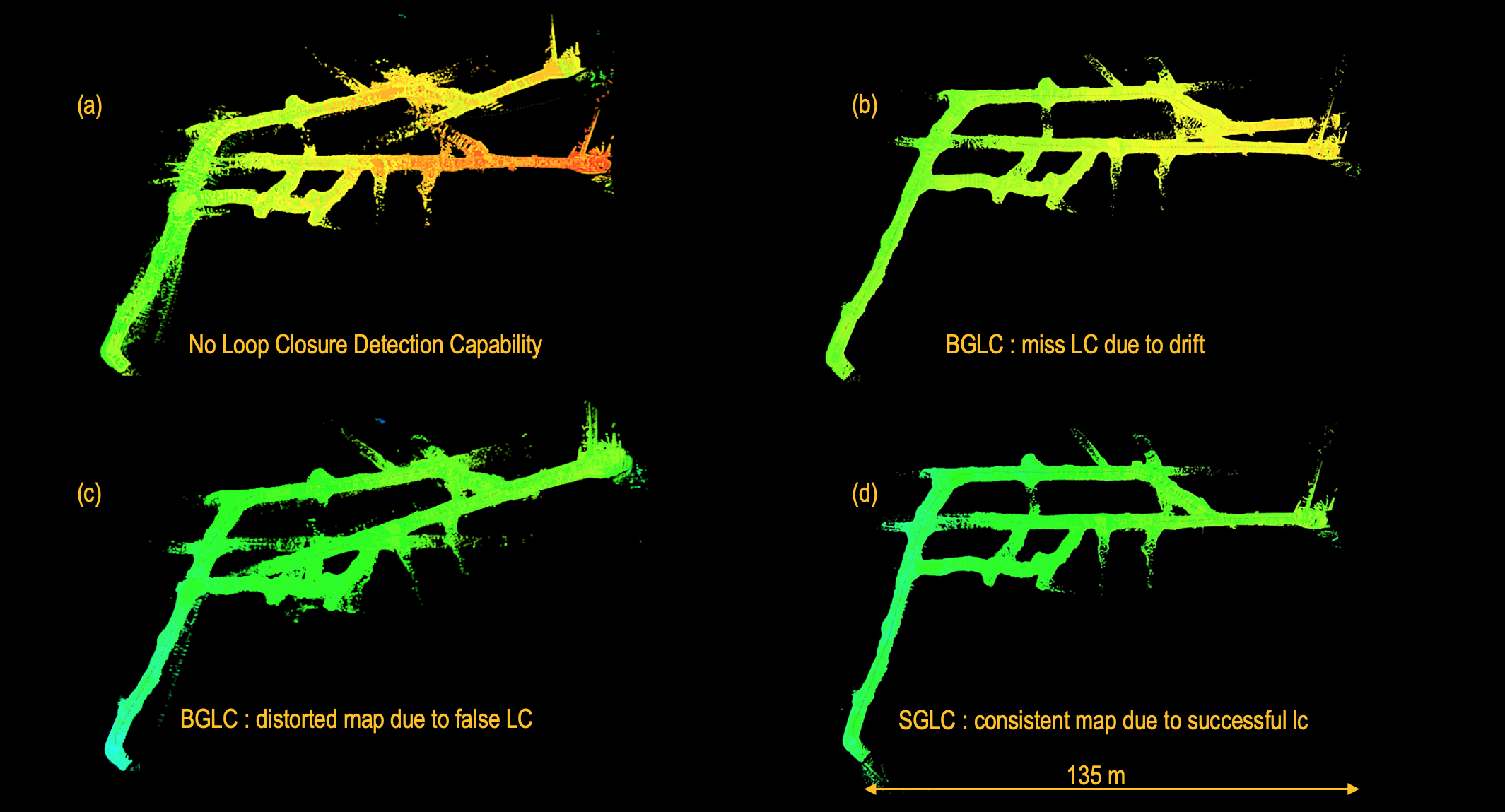}
  \caption{Top-down view of the 3D Map of the Beckley coal mine, Beckley, WV. (a) mapping with no loop closure detection capability. (b) and (c) the BGLC method maps results in distorted maps due to spurious and missed loop closure opportunities. (d) map using the proposed SGLC method. \label{fig:beckley_lc}}
\end{figure}\begin{figure}[b!]
	\includegraphics[width=1.0\columnwidth]{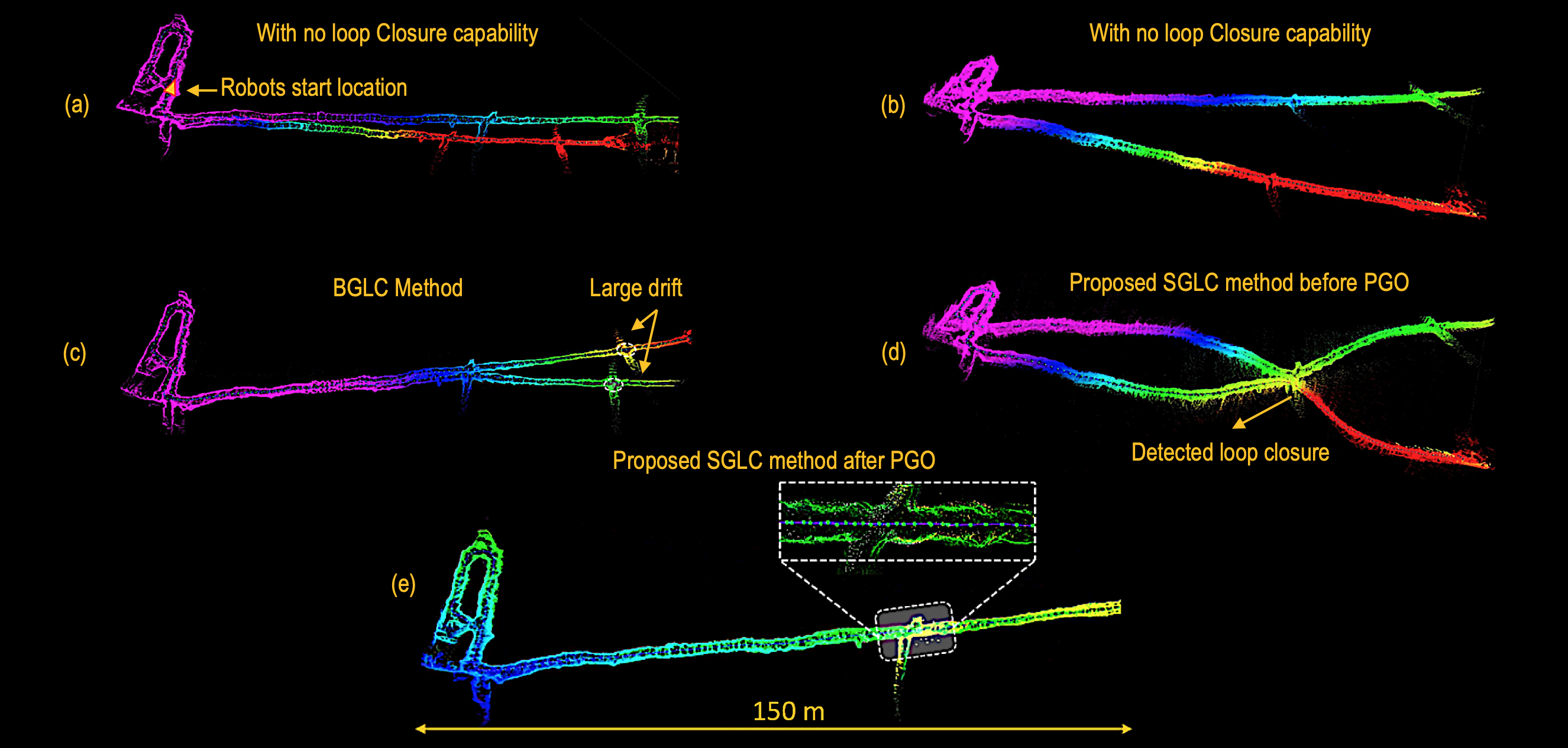}
	\caption{Map of the Eagle Mine, Julian, CA obtained on a base station. (a) and (b) present the top-down and side views of the 3D map obtained on the base station with no loop closure detection capability. (c) mapping result using the BGLC method. (d) mapping results using the proposed SGLC method, before performing PGO to show the detected loop closure between two nodes in the graphs . (e) global map after performing PGO using the proposed SGLC method.}
	\label{fig:lc_eagle_mine}
\end{figure}

Fig. \ref{fig:beckley_lc} and Fig. \ref{fig:lc_eagle_mine} show examples of constructed maps in two underground mines. At the Beckley Coal mine a robot autonomously traverses more than $1$km in the mine and returns to the start location to close the loop. Fig. \ref{fig:beckley_lc}-(a) shows the constructed map in an open-loop scenario where no loop closure detection capability is used. Fig. \ref{fig:beckley_lc}-(b-c) show the maps obtained in two trials using the BGLC method where a fixed search radius of $10$m is used. The results show many loop closure opportunities are missed, leading to the dramatic distortion of the maps. Fig. \ref{fig:beckley_lc}-(d) shows the map obtained using our proposed SGLC method. As the method is pose-invariant, it is unaffected by accumulation of errors in robot trajectory which results in successful detection of loop closures and subsequently a more consistent representation of the environment.

Fig. \ref{fig:lc_eagle_mine} presents a multi-robot collaborative mapping scenario where local maps obtained by a team of two robots deployed in the Eagle mine are merged on a base station.
Both robots with known initial poses in the world coordinate system autonomously navigate $500$m inside the tunnels of the mine. Fig. \ref{fig:lc_eagle_mine} presents the top (a) and side (b) views of the maps obtained on the base station in an open loop scenario, without using any loop closure detection capability. As the error in robot trajectories accumulates over time, the maps start to drift apart unbounded in the absence of loop closure detections.
Using the BGLC method, for every key-node in one graph, key-nodes in the other graph that lie inside a search radius of $10$m are considered for loop closure.
As presented in Fig. \ref{fig:lc_eagle_mine}-(c), while the method performs well when the drift in robot trajectories is small, it misses many loop closure opportunities when the trajectories drift apart due to accumulation of noisy odometric estimates.
Fig. \ref{fig:lc_eagle_mine}-(d) shows a successful loop closure using the proposed SGLC method before performing pose graph optimization. Using a salient geometric feature (i.e., a T-junction in the tunnel) an inter-robot loop closure is detected and the diverged trajectories are joined again. Fig. \ref{fig:lc_eagle_mine}-(e), shows a consistent map of the environment after pose graph optimization and minimization of the error.
\begin{figure}[b!]   
\centering 
  \includegraphics[width=1.0\columnwidth]{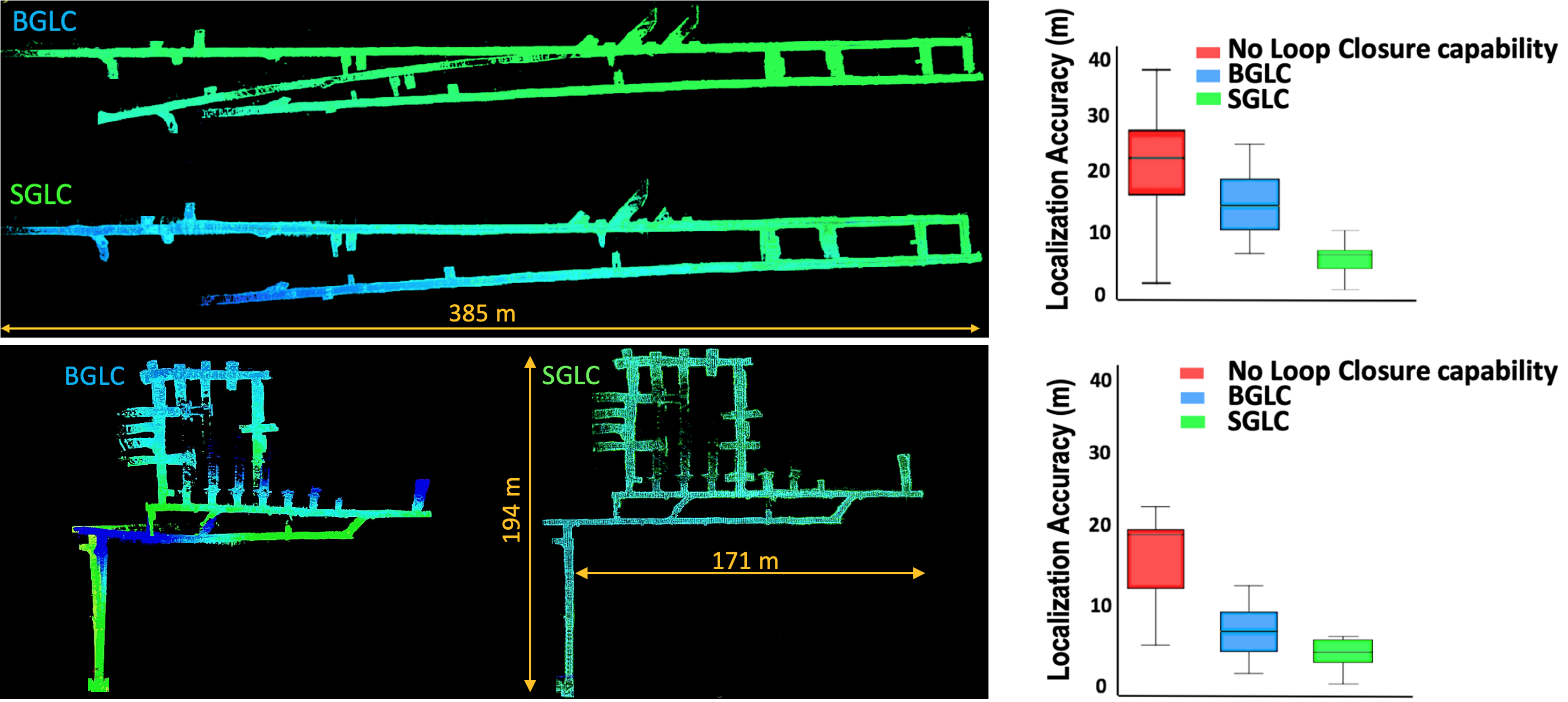}
  \caption{Maps of the Bruceton Experimental and Safety Research mines constructed using the BGLC and proposed SGLC methods. The box plots report the object localization errors in the final maps, without loop closure detection capability, as well as the BGLC and the proposed SGLC methods. \label{fig:EX-SRCM_lc}}
\end{figure}
Fig. \ref{fig:EX-SRCM_lc} reports examples of multi-robot mapping for a team of two robots deployed in the Bruceton Experimental Mine and Safety Research Coal Mine during the Tunnel Circuit of DARPA Subterranean Challenge\cite{SubT}, in August of 2019. 
Using the BGLC method, large drifts are visible in both maps as the base station fails to detect inter-robot loop closures to merge the maps due to many missed loop closure opportunities when the maps start to drift apart. The maps obtained using the proposed SGLC method show a more consistent 3D representation of the mines due to more frequent loop closures that help reduce the drift in the estimated trajectories.

At the Tunnel Circuit of the DARPA Subterranean challenge, a variety of objects (e.g., backpack, fire extinguisher, drill, survivor, and a cellphone) were placed at unknown locations in the tunnels of the mines. 
The autonomous robots were used to detect and localize these objects in the environment by relying on the onboard RGB-D camera and the lidar-based SLAM. We use the estimated location of the detected objects to provide a quantitative evaluation of localization and mapping accuracy. The box plots in Fig. \ref{fig:EX-SRCM_lc} report a quantitative evaluation of localization accuracy using no loop closure detection capability, the BGLC method, and the proposed SGLC method by comparing the estimated location of detected objects against the ground truth data provided by DARPA.
With no loop closure detection capability, the robot pose uncertainty and location uncertainty of detected objects grow as the errors in lidar odometry accumulate. By using the BGLC method the drift is reduced as compared to the open loop trajectory, but several missed loop closure opportunities lead to accumulation of errors in robot trajectories which manifests itself as large object localization errors. Using the proposed SGLC method, loop closures are consistently detected as the robot navigates the unknown environment, which results in significantly more accurate localization and mapping results.
\begin{figure}[t!]   
\centering 
  \includegraphics[width=1.0\columnwidth]{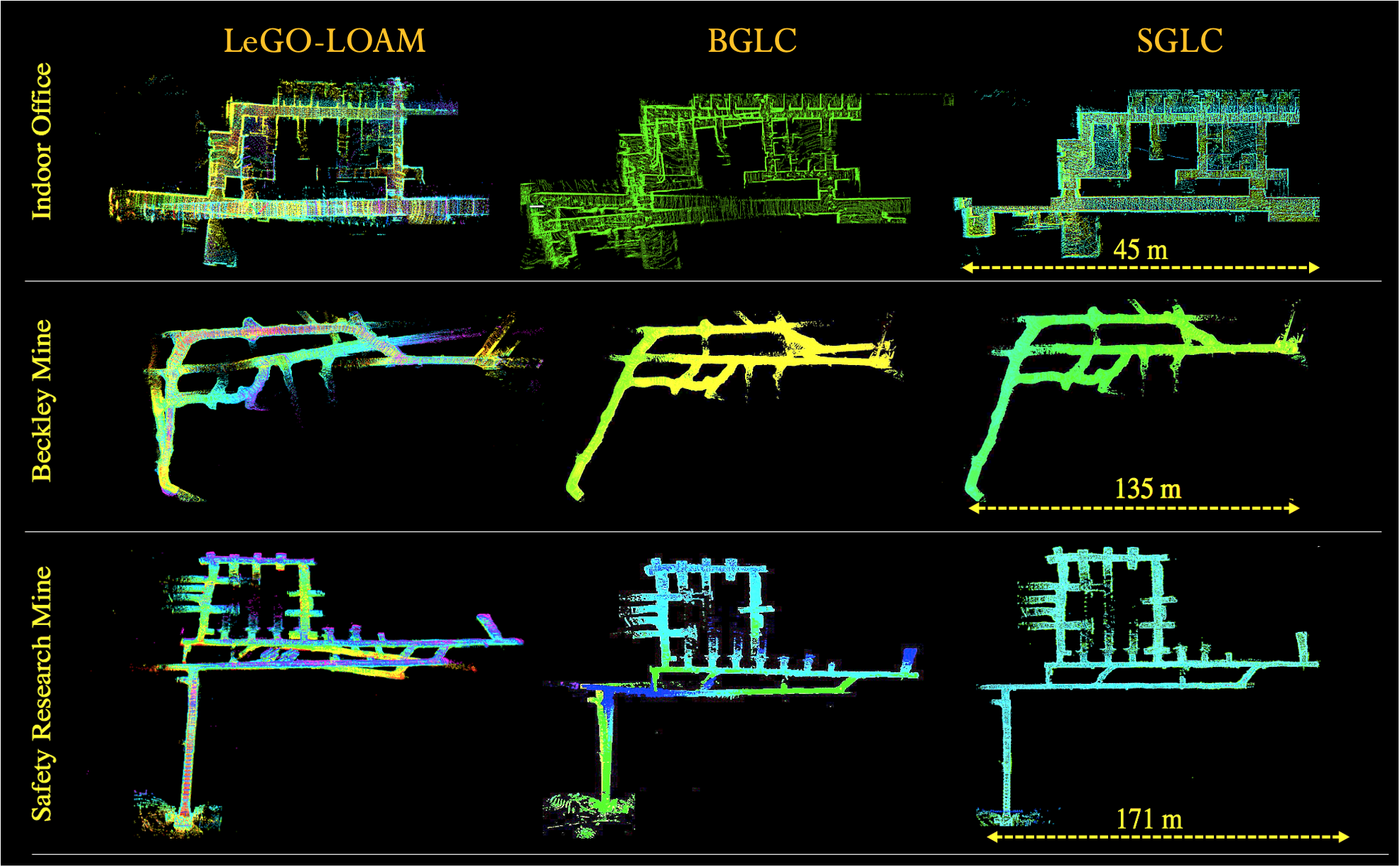}
  \caption{Maps of the indoor office, Beckley Coal mine, and Safety Research mine, using the LeGO-LOAM, BGLC, and proposed SGLC methods. \label{fig:map_comparisons1}}
\end{figure}
\begin{figure}[t!]   
\centering 
  \includegraphics[width=1.0\columnwidth]{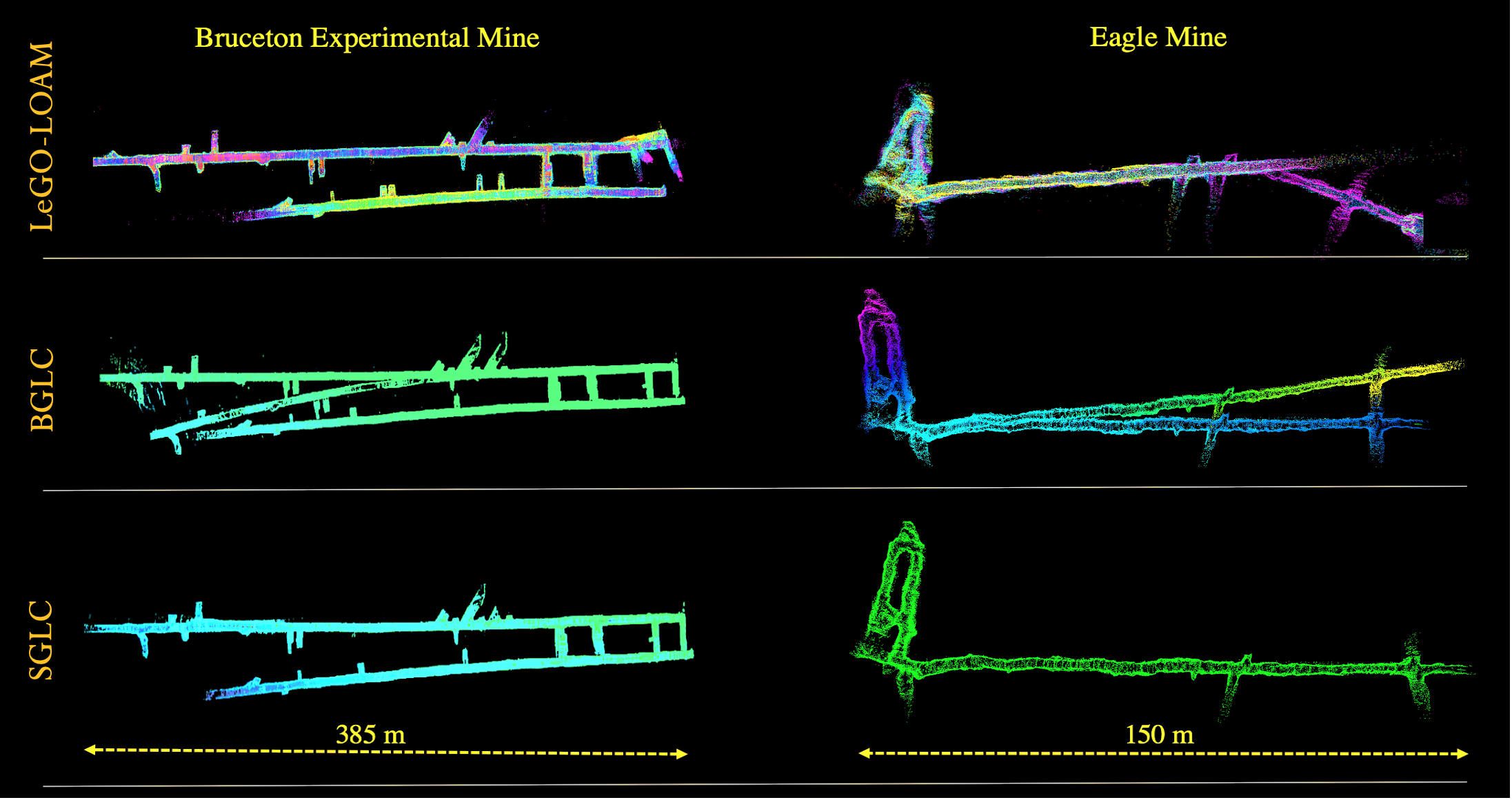}
  \caption{Maps of the Experimental mine, and Eagle mine using the LeGO-LOAM, BGLC, and proposed SGLC methods \label{fig:map_comparisons2}}
\end{figure}
\begin{figure}[t!]
\centering 
  \includegraphics[width=1.0\columnwidth]{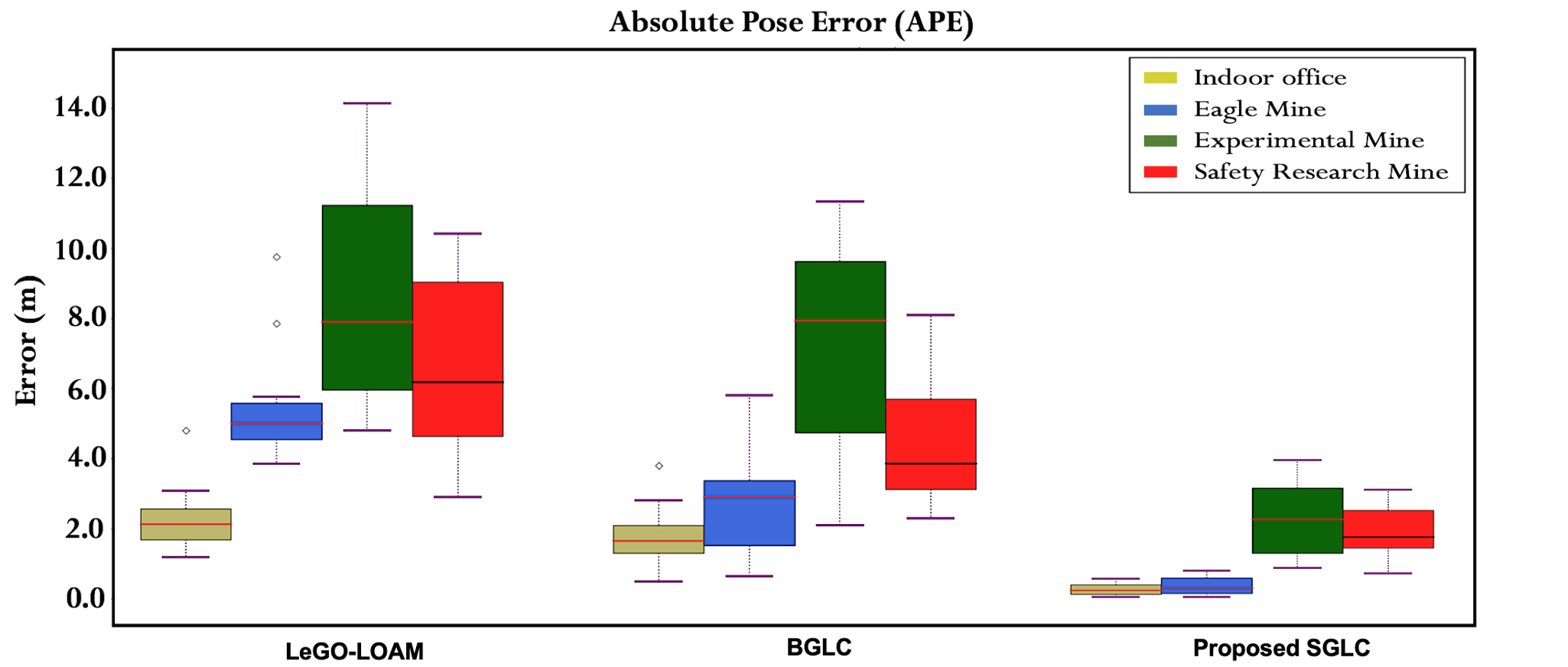}
  \caption{Quantitative comparison of mapping results based on the Absolute Pose Error (APE) using LeGO-LOAM, BGLC, and the proposed SGLC methods in $4$ perceptually-degraded and large-scale subterranean environments. The box plots are obtained by running each algorithm $10$ times on the dataset collected in each environment and measuring the absolute pose error. \label{fig:map_comparisons_quant}}
\end{figure}
Fig. \ref{fig:map_comparisons1} and Fig. \ref{fig:map_comparisons2} provide comparison of the 3D maps obtained using the LeGO-LAOM, BGLC and proposed SGLC method in five perceptually-degraded and large-scale underground environments. While in LeGO-LOAM the default search radius for loop closures is set to $7$m, in our experiments the search radius was increased to $10$m for both LeGO-LOAM and the BGLC methods to improve loop closure detections in presence of large drift. The results show both LeGO-LOAM and BGLC methods initially perform well as long as the drift in robot trajectories is small, but missing several loop closure opportunities in the presence of large drift. This exposes the vulnerability of pose-dependent loop closure detection methods, especially in large-scale and perceptually-degraded environments where accumulation of odometry errors can be significant. The results show the proposed SGLC method consistently outperforms the LeGO-LOAM and BGLC methods due to its drift-resiliency.

Fig. \ref{fig:map_comparisons_quant} reports comparison of mapping results based on the Absolute Pose Error (APE) metric for LeGO-LOAM, BGLC, and the proposed SGLC methods using datasets obtained from four perceptually-degraded and large-scale subterranean environments. The box plots are obtained by running each algorithm $10$ times on each dataset, and recording the localization accuracy of several known landmarks along the robot trajectory. The results show the SGLC method consistently outperforms LeGO-LOAM and BGLC methods across all environments with different levels of perceptual degradation, and thus, the drift in estimated robot trajectories is significantly reduced as reflected in small APE values. The results also show all three methods suffer from relatively larger drift in the Experimental and Safety Research mines. This is due to presence of several long, flat and featureless corridors in these mines which leads to more noisy odometric estimates.

\section{Conclusion and Future Work} \label{sec:discussion}
While dramatic progress has been made over the past few decades in the field of single- and multi-robot SLAM, localization and mapping in unknown, large-scale, and perceptually-degraded environments presents a variety of challenges due to sparsity of distinctive landmarks, self-similarity of features, and poor illumination.
In these extreme settings, the front-end is exposed to major challenges including data association ambiguity that can result in significant drift in the estimated robot trajectory during long-term or large-scale navigation. This highlights the importance of drift-resilient and robust loop closure detection methods that can reduce the drift, and also enable merging individual maps obtained by a team of robots into a globally consistent map of the environment in a multi-robot SLAM system.

In this paper, we developed a set of metrics and methods to improve detection of intra- and inter-robot loop closures to increase robustness and reliability of autonomous navigation in unknown and extreme subterranean environments.
We developed a degeneracy-aware SLAM front-end that is capable of determining the level of geometric degeneracy through eigenanalysis of the solution of the ICP algorithm. This was validated by comparing the results obtained from lidar odometry against the wheel-inertial odometry in cases where the wheel odometry was known to be reliable.
While most state-of-the-art methods attempt loop closures periodically without considering the level of geometric degeneracy of the scene, it is crucial to avoid closing loops in ambiguous areas with high level of geometric degeneracy as it could result in catastrophic distortions of the map.
Using our proposed degeneracy-aware front-end, areas with high level of geometric degeneracy that can lead to data association ambiguity and spurious loop closures are removed from loop closure consideration. In contrast to traditional loop closure detection methods that perform the search for loop closures only locally, this capability allows the search for loop closures to be expanded as needed to account for drift, without increasing the probability of spurious loop closures. Through analysis of the geometric degeneracy metric $\kappa$, in a verity of indoor and underground environments, it was shown the metric can be reliably used to determine geometric degeneracy in unknown environments.

Moreover, a drift-resilient geometric loop closing method based on saliency maps was proposed to enable a multi-stage place recognition pipeline through which loop closure candidates can be identified over the entire robot trajectory independent of estimated robot pose. This increases the robustness and accuracy of loop closure detections, especially in large-scale and perceptually-degraded environments.
Putative loop closures were identified by evaluating the similarity of the spatial configuration of the scenes using a feature-based image registration process, before being validated using geometric and outlier rejection steps.
Through extensive real-world experiments and comparisons with state-of-the-art methods, it was shown the proposed method improves localization and mapping accuracy in perceptually-degraded subterranean environments where commonly used loop closure detection methods exhibit insufficient robustness and accuracy.

Future work will focus on developing a robust multi-sensor lidar-centric front-end using the geometric degeneracy detection capability.
While in this work the determination of geometric degeneracy was only used to constrain the search for loop closures to the most observable areas in an unknown environment, the same capability can be used to enable a robust multi-sensor degeneracy-aware odometry system where the front-end can evaluate the reliability of odometric estimates in real-time and switch to alternative odometry sources where the unknown environment is determined to be geometrically degenerate.
Another direction for future research is extending this work to a distributed multi-robot SLAM architecture.
While the centralized multi-robot architecture presented in this paper works well for a small team of robots where a base station is in charge of map alignment and merging, in large-scale underground environments with sporadic communication with the base station, a distributed collaborative architecture is desirable to enable a team of robots to exchange their maps in order to maintain a consistent global map of the explored environment.


\begin{thebibliography}{}
\bibitem{SPOT} A. Bouman, M. F. Ginting, N. Alatur, M. Palieri, D. D. Fan, T. Touma, T. Pailevanian, S. K. Kim, K. Otsu, J. Burdick, A. Agha-Mohammadi. ``Autonomous Spot: Long-range Autonomous Exploration of  Extreme Environments with Legged Locomotion.'' IEEE/RSJ International Conference on Intelligent Robots and Systems, Las Vegas, NV, 2020.
\bibitem{Choudhary} S. Choudhary, L. Carlone, C. Nieto, J. Rogers, Z. Liu, H. I. Christensen, and F. Dellaert. ``Multi robot object-based slam. '' International Symposium on Experimental Robotics, pp. 729-741. Springer, Cham, 2016.
\bibitem{Carlone3} S. Choudhary, L. Carlone, C. Nieto, J. Rogers, H. I. Christensen, and F. Dellaert. ``Distributed mapping with privacy and communication constraints: Lightweight algorithms and object-based models.'' The International Journal of Robotics Research, vol. 36, no. 12, 1286-1311. 2017.
\bibitem{Cadena} C. Cadena, L. Carlone, H. Carrillo, Y. Latif, D. Scaramuzza, J. Neira, I. Reid, and J. J. Leonard. ``Past, present, and future of simultaneous localization and mapping: Toward the robust-perception age.'' IEEE Transactions on robotics, vol. 32, no. 6, pp. 1309-1332, 2016.
\bibitem{AGUFall2019} A. Agha, K. L. Mitchell, and P. J. Boston. ``Robotic exploration of planetary subsurface voids in search for life.'' In AGU Fall Meeting Abstracts, vol. 2019, pp. P41C-3463. 2019.
\bibitem{Haruyama} J. Haruyama, T. Morota, S. Kobayashi, S. Sawai, P. G. Lucey, M. Shirao, and M. N. Nishino. ``Lunar Holes and Lava Tubes as Resources for Lunar Science and Exploration.'' Moon, pp. 139–163, 2012.
\bibitem{ORB-SLAM} R. Mur-Artal, J. M. M. Montiel, and J. D. Tardos. ``ORB-SLAM: a versatile and accurate monocular SLAM system.'' IEEE transactions on robotics vol. 31, no. 5, pp. 1147-1163, 2015.
\bibitem{LOAM} J. Zhang, and S. Singh. ``LOAM: Lidar Odometry and Mapping in Real-time.'' Robotics: Science and Systems, vol. 2, p. 9. 2014.
\bibitem{Bloesch} M. Bloesch, S. Omari, M. Hutter, and R. Siegwart. ``Robust visual inertial odometry using a direct EKF-based approach.'' International Conference on Intelligent Robots and Systems, IEEE, 2015, pp. 298–304.
\bibitem{Leutenegger} S. Leutenegger, S. Lynen, M. Bosse, R. Siegwart, and P. Furgale. ``Keyframe-based visual–inertial odometry using nonlinear optimization.'' The International Journal of Robotics Research, vol. 34, no. 3, pp. 314–334, 2015.
\bibitem{Khattak1} S. Khattak, F. Mascarich, T. Dang, C. Papachristos, and K. Alexis. ``Robust Thermal-Inertial Localization for Aerial Robots: A Case for Direct Methods.'' International Conference on Unmanned Aircraft Systems, pp. 1061-1068. IEEE, 2019.
\bibitem{LION} A. Tagliabue, J. Tordesillas, X. Cai, A. S. Navarro, and J. P. How, L. Carlone, and A. Agha-Mohammadi. ``LION: Lidar-Inertial Observability-Aware Navigator for Vision-Denied Environments'', International Symposium on Experimental Robotics 2021
\bibitem{UWB} N. Funabiki, B. Morrell, J. Nash, and A. Agha-mohammadi. ``Range-Aided Pose-Graph-Based SLAM: Applications of Deployable Ranging Beacons for Unknown Environment Exploration.'' IEEE Robotics and Automation Letters, vol. 6, no. 1, 48-55
\bibitem{LOCUS} M. Palieri, B. Morrell, A. Thakur, K. Ebadi, J. Nash, A. Chatterjee, C. Kanellakis, L. Carlone, C. Guaragnella, and A. Agha-mohammadi. ``LOCUS: A Multi-Sensor Lidar-Centric Solution for High-Precision Odometry and 3D Mapping in Real-Time.'' IEEE Robotics and Automation Letters, vol. 6, no. 2 (2020): 421-428. 
\bibitem{HERO} A. Santamaria-navarro, R. Thakker, D. D. Fan, B. Morrell, and A. Agha-mohammadi. ``Towards resilient autonomous navigation of drones.'' International Symposium on Robotics Research, 2019.
\bibitem{vision-based-navigation-survey} S. Lowry, N. Sünderhauf, P. Newman, J. J. Leonard, D. Cox, P. Corke, and M. J. Milford. ``Visual place recognition: A survey.'' IEEE Transactions on Robotics, vol. 32, no. 1, 1-19. 2015.
\bibitem{ORBSLAM} R. M. Artal, J. M. Montiel, and J. D. Tardos. ``ORB-SLAM: a versatile and accurate monocular SLAM system.'' IEEE transactions on robotics, vol. 31, no. 5, 1147-1163. 2015.
\bibitem{Lopez} D. G. Lopez, and J. D. Tardos. ``Real-time loop detection with bags of binary words.'' IEEE International Conference on Intelligent Robots and Systems, pp. 51-58. 2011.
\bibitem{Kamak1} K. Ebadi, and A. Agha-Mohammadi. ``Rover Localization in Mars Helicopter Aerial Maps: Experimental Results in a Mars-Analogue Environment.'' In International Symposium on Experimental Robotics, pp. 72-84. Springer, Cham, 2018.
\bibitem{Kamak2} K. Ebadi, and S. Wood. ``Scene Matching-based Localization of Unmanned Aerial Vehicles in Unstructured Environments.'' In 2018 52nd Asilomar Conference on Signals, Systems, and Computers, pp. 1519-1523. IEEE, 2018.
\bibitem{vision-lidar-lc} P. Newman, G. Sibley, M. Smith, M. Cummins, A. Harrison, C. Mei, I. Posner et al. ``Navigating, recognizing and describing urban spaces with vision and lasers.'' The International Journal of Robotics Research, vol. 28, no. 11-12, 1406-1433. 2009
\bibitem{vision-lidar-lc2} H. Qin, M. Huang, J. Cao, and X. Zhang. ``Loop closure detection in SLAM by combining visual CNN features and submaps.'' In 2018 4th International Conference on Control, Automation and Robotics (ICCAR), pp. 426-430. IEEE, 2018.
\bibitem{Nuchter} A. Nuchter, H. Surmann, K. Lingemann, J. Hertzberg, and S. Thrun. ``6D SLAM with an application in autonomous mine mapping.'' IEEE International Conference on Robotics and Automation, vol. 2, pp. 1998-2003. IEEE, 2004.
\bibitem{LOAM2} J. Zhang, and S. Singh. ``Low-drift and real-time lidar odometry and mapping.'' Autonomous Robots 41, no. 2 (2017): 401-416.
\bibitem{LLOAM} X. Ji, L. Zuo, C. Zhang, and Y. Liu. ``LLOAM: LiDAR Odometry and Mapping with Loop-closure Detection Based Correction.'' IEEE International Conference on Mechatronics and Automation, pp. 2475-2480, 2019.
\bibitem{SegMatch} R. Dubé, D. Dugas, E. Stumm, J. Nieto, R. Siegwart, and C. Cadena. ``Segmatch: Segment based place recognition in 3d point clouds.'' IEEE International Conference on Robotics and Automation, pp. 5266-5272. IEEE, 2017.
\bibitem{SegMap} R. Dubé, A. Cramariuc, D. Dugas, J. Nieto, R. Siegwart, and C. Cadena. ``SegMap: 3D segment mapping using data-driven descriptors.'' arXiv preprint arXiv:1804.09557, 2018.
\bibitem{LeGoLOAM} T. Shan, and B. Englot. ``LeGO-LOAM: Lightweight and ground-optimized lidar odometry and mapping on variable terrain.'' IEEE International Conference on Intelligent Robots and Systems, pp. 4758-4765. IEEE, 2018.
\bibitem{Cartographer} W. Hess, D. Kohler, H. Rapp, and D. Andor. ``Real-time loop closure in 2D LIDAR SLAM.'' IEEE International Conference on Robotics and Automation, pp. 1271-1278. IEEE, 2016.
\bibitem{SPA} K. Konolige, G. Grisetti, R. Kummerle, W. Burgard, B. Limketkai, and R. Vincent, ``Sparse pose adjustment for 2D mapping.'' IEEE International Conference on Intelligent Robots and Systems, Taipei, Taiwan, 2010.
\bibitem{Thrun} S. Thrun, D. Hahnel, D. Ferguson, M. Montemerlo, R. Triebel, W. Burgard, C. Baker, Z. Omohundro, S. Thayer, and W. Whittaker. ``A system for volumetric robotic mapping of abandoned mines.'' IEEE International Conference on Robotics and Automation, vol. 3, pp. 4270-4275. 2003.
\bibitem{ICP} C. Yang, and G. G. Medioni. ``Object modeling by registration of multiple range images.'' Image and Vision Computing, vol. 10, no. 3, pp. 145-155. 1992.
\bibitem{Tardioli2} D. Tardioli, D. Sicignano, L. Riazuelo, J. L. Villarroel, and L. Montano. ``Robot teams for exploration in underground environments.'' In Workshop ROBOT11: Robotica Experimental, pp. 205-212. 2012.
\bibitem{Tardioli3} D. Tardioli, L. Riazuelo, T. Seco, J. Espelosín, J. Lalana, J. L. Villarroel, and L. Montano. ``A robotized dumper for debris removal in tunnels under construction.'' In Iberian Robotics conference, Springer, Cham, pp. 126-139. 2017.
\bibitem{Zlot} R. Zlot and M. Bosse. ``Efficient large-scale 3d mobile mapping and surface reconstruction of an underground mine.'' In Field and service robotics, Springer, Berlin, Heidelberg, pp. 479-493. 2014.
\bibitem{Leingartner} M. Leingartner, J. Maurer, A. Ferrein, and G. Steinbauer. ``Evaluation of sensors and mapping approaches for disasters in tunnels.'' Journal of field robotics vol. 33, no. 8, pp. 1037-1057, 2016.
\bibitem{Jacobson} A. Jacobson, F. Zeng, D. Smith, N. Boswell, T. Peynot, and M. Milford. ``Semi-supervised slam: Leveraging low-cost sensors on underground autonomous vehicles for position tracking.'' IEEE International Conference on Intelligent Robots and Systems, pp. 3970-3977. 2018.
\bibitem{Cox} Cox, Ingemar J. ``Blanche-an experiment in guidance and navigation of an autonomous robot vehicle.'' IEEE Transactions on robotics and automation, vol. 7, no. 2, pp. 193-204. 1991.
\bibitem{Gutmann} J. S. Gutmann and C. Schlegel. Amos. ``Comparison of scan matching approaches for self- localization in indoor environments.'' IEEE Proceedings of the First Euromicro Workshop on Advanced Mobile Robot, page 61–67, 1996.
\bibitem{FLU2} F. Lu. “Shape registration using optimization for mobile robot navigation.'' Ph.D. thesis, University of Toronto, 1995.
\bibitem{Harris} C.  Harris  and  M.J.  Stephens. ``A  combined  corner  and  edge  detector.'' Alvey Vision Conference, pages 147–152. 1988.
\bibitem{fpfh} R. B. Rusu, N. Blodow, and M. Beetz. ``Fast point feature histograms (FPFH) for 3D registration.'' In IEEE International Conference on Robotics and Automation, pp. 3212-3217. 2009.
\bibitem{LiOlson} Y. Li and E. Olson. ``Extracting general-purpose features from lidar data.'' IEEE International Conference on Robotics and Automation, pp. 1388-1393. 2010 
\bibitem{Tipaldi} G. D. Tipaldi, L. Spinello, and W. Burgard.  ``Geometrical flirt phrases for large scale  place  recognition  in  2d  range  data.'' IEEE  International  Conference  on Robotics and Automation, pp. 2693–2698. 2013.
\bibitem{Bosse} Bosse, Michael, and Robert Zlot. ``Continuous 3D scan-matching with a spinning 2D laser.'' In 2009 IEEE International Conference on Robotics and Automation, pp. 4312-4319. 2009.
\bibitem{Dorit} D. Borrmann, J. Elseberg, K. Lingemann, A. Nüchter, and J. Hertzberg. ``Globally consistent 3D mapping with scan matching.'' Robotics and Autonomous Systems, vol. 56, no. 2, 130-142. 2008.
\bibitem{Bing} B. J. Ho, , P. Sodhi, P. Teixeira, M. Hsiao, T. Kusnur, and M. Kaess. ``Virtual occupancy grid map for submap-based pose graph SLAM and planning in 3D environments.'' In International Conference on Intelligent Robots and Systems (IROS), pp. 2175-2182. IEEE, 2018.
\bibitem{LAMP} K. Ebadi, Y. Change, M. Palieri, A. Stephens, A. H. Hatteland, E. Heiden, A. Thakur, B. Morrell, L. Carlone, A. Agha-mohammadi. ``LAMP: Large-scale autonomous mapping and positioning for exploration of perceptually-degraded subterranean environments.'' In 2020 IEEE International Conference on Robotics and Automation (ICRA), pp. 80-86. 2020.
\bibitem{SubT} "DARPA Subteranean (SubT) Challenge.'' [Online]. Available: https://www.subtchallenge.com
\bibitem{Random_filter} Z. J. Yew, and G. H. Lee. ``3DFeat-Net: Weakly supervised local 3D features for point cloud registration.'' In European Conference on Computer Vision, Springer, Cham, pp. 630-646. 2018.
\bibitem{pointCloud_filter} X. F. Han, J. S. Jin, M. J. Wang, W. Jiang, L. Gao, and L. Xiao. ``A review of algorithms for filtering the 3D point cloud.'' Signal Processing: Image Communication, no. 57, 103-112. 2017
\bibitem{GICP} A. Segal, D. Haehnel, and S. Thrun. ``Generalized-ICP.'' In Robotics: science and systems, vol. 2, no. 4, p. 435. 2009.
\bibitem{Ila} V. Ila, J. M. Porta, and J. Andrade-Cetto, ``Information-based compact pose SLAM.'' IEEE Transactions on Robotics, vol. 26, no. 1, pp. 78–93, 2010.
\bibitem{Johannsson} H. Johannsson, M. Kaess, M. Fallon, and J. J. Leonard. ``Temporally scalable visual SLAM using a reduced pose graph.'' In IEEE International Conference on Robotics and Automation, pp. 54-61. 2013.
\bibitem{Kretzschmar} H. Kretzschmar, C. Stachniss, and G. Grisetti, ``Efficient information theoretic graph pruning for graph-based SLAM with laser range finders.'' In IEEE International Conference on Intelligent Robots and Systems, pp. 865–871. 2011.
\bibitem{Szeliski} R. Szeliski. ``Computer vision algorithms and applications.'' Springer Science and Business Media, 2010.
\bibitem{condition_number} J. Zhang, M. Kaess, and S. Singh. ``On degeneracy of optimization-based state estimation problems.'' In IEEE International Conference on Robotics and Automation (ICRA), pp. 809-816. 2016.
\bibitem{ICPcomplexity}Jost, Timothée, and Heinz Hügli. ``Fast ICP algorithms for shape registration.'' In Joint Pattern Recognition Symposium, Springer, Berlin, Heidelberg, pp. 91-99. 2002.
\bibitem{FMcomplexity} M. Toews, C. Wachinger, R. S. J. Estepar, and W. M. Wells. ``A feature-based approach to big data analysis of medical images.'' In International Conference on Information Processing in Medical Imaging, Springer, Cham, pp. 339-350. 2015.
\bibitem{local_minima} N. Gelfand, L. Ikemoto, S. Rusinkiewicz, and M. Levoy. ``Geometrically stable sampling for the ICP algorithm.'' In Fourth International Conference on 3D Digital Imaging and Modeling, Proceedings., pp. 260-267. IEEE, 2003.
\bibitem{Elfes} A. Elfes ``Using occupancy grids for mobile robot perception and navigation.'' Computer, vol. 22, no. 6, 46-57. 1989.
\bibitem{ConfidenceRichMapping} A. Agha-mohammadi, E. Heiden, K. Hausman, and G. Sukhatme. ``Confidence-rich grid mapping.'' The International Journal of Robotics Research, vol. 38, no. 12-13, pp. 1352–1374, 2019.
\bibitem{Costmap} [Online]. Available: https://www.wiki.ros.org/costmap\_2d/hydro/staticmap
\bibitem{ORB} E. Rublee, V. Rabaud, K. Konolige, and G. Bradski. ``ORB: An efficient alternative to SIFT or SURF.'' In International Conference on Computer Vision, pp. 2564-2571. IEEE, 2011.
\bibitem{Silpa} S. Chanop, and R. Hartley. ``Optimised KD-trees for fast image descriptor matching.'' IEEE Conference on Computer Vision and Pattern Recognition, 2008.
\bibitem{RANSAC} M. A. Fischler, and R. C. Bolles. ``Random sample consensus: A paradigm for model fitting with applications to image analysis and automated cartography.'' Communications of the ACM, vol. 24, no. 6, 381-395. 1981.
\bibitem{RANSAC_outlier_rejection} B. Ruzgiene and W. Forstner .Ruzgiene, Birute, and Wolfgang Förstner. ``Ransac for outlier detection.'' Geodezija ir kartografija, vol. 31, no. 3, 83-87. 2005.
\bibitem{YOLO} J. Redmon, and A. Farhadi. ``Yolov3: An incremental improvement.'' arXiv preprint arXiv:1804.02767 (2018). 
\bibitem{Apriltag3} J. Wang and E. Olson, ``AprilTag 2: Efficient and robust fiducial detection.'' IEEE International Conference on Intelligent Robots and Systems, 2016.
\bibitem{GTSAM} F. Dellaert, ``Factor graphs and GTSAM: A hands-on introduction.'' Georgia Institute of Technology, Atlanta, GA, Tech. Rep.
\bibitem{Door_SLAM} P. Y. Lajoie, B. Ramtoula, Y. Chang, L. Carlone, and G. Beltrame. ``DOOR-SLAM: Distributed, online, and outlier resilient SLAM for robotic teams.'' IEEE Robotics and Automation Letters, vol. 5, no. 2, 1656-1663. 2020.
\bibitem{PCM} J. G. Mangelson, D. Dominic, R. M. Eustice, and R. Vasudevan. ``Pairwise consistent measurement set maximization for robust multi robot map merging.'' In IEEE International Conference on Robotics and Automation, pp. 2916-2923. 2018.
\bibitem{Carlone14} L. Carlone, R. Aragues, J. A. Castellanos, and B. Bona. ``A fast and accurate approximation for planar pose graph optimization.'' International Journal of Robotics Research, vol. 33, no.7, pp. 965–987. 2014.
\bibitem{PUCK} ``Velodyne Puck LITE lidar.'' [Online]. Available: https://velodynelidar.com/vlp-16-lite.html
\bibitem{Grupp} M. Grupp, ``evo: Python package for the evaluation of odometry and SLAM.'' [Online]. Available: https://github.com/MichaelGrupp/evo, 2017.
\end{thebibliography}
\end{document}